\documentclass[10pt,twocolumn,letterpaper]{article}

\usepackage{cvpr}              
\definecolor{cvprblue}{rgb}{0.21,0.49,0.74}
\usepackage[pagebackref,breaklinks,colorlinks,allcolors=cvprblue]{hyperref}

\usepackage{enumitem}
\usepackage{threeparttable}
\usepackage{graphicx,verbatim}
\usepackage{booktabs} 
\usepackage{multirow}
\usepackage{bbm} 
\usepackage{amssymb}
\usepackage{wrapfig} 
\usepackage{array}
\usepackage{graphicx}
\usepackage{subcaption}
\usepackage{bm}
\usepackage{amsmath}
\usepackage{hyperref}
\usepackage{grffile}
\usepackage{pifont} 
\usepackage{mathtools}
\usepackage{tabularx}
\usepackage{float}
\usepackage{booktabs} 
\usepackage{color}
\usepackage{adjustbox} 
\usepackage[table]{xcolor}
\usepackage{titletoc}
\usepackage{tcolorbox}
\tcbuselibrary{breakable}
\usepackage{bbding}
\usepackage{soul}
\usepackage{ulem}

\def\paperID{N/A} 
\def\confName{CVPR}
\def\confYear{2026}

\newcommand{\ourbench}{{\fontfamily{ppl}\selectfont OmniBrainBench}}
\newcommand{\ourvqa}{{\fontfamily{ppl}\selectfont OmniBrainVQA}}

\title{\raisebox{-0.3\height}{\includegraphics[width=1.28cm]{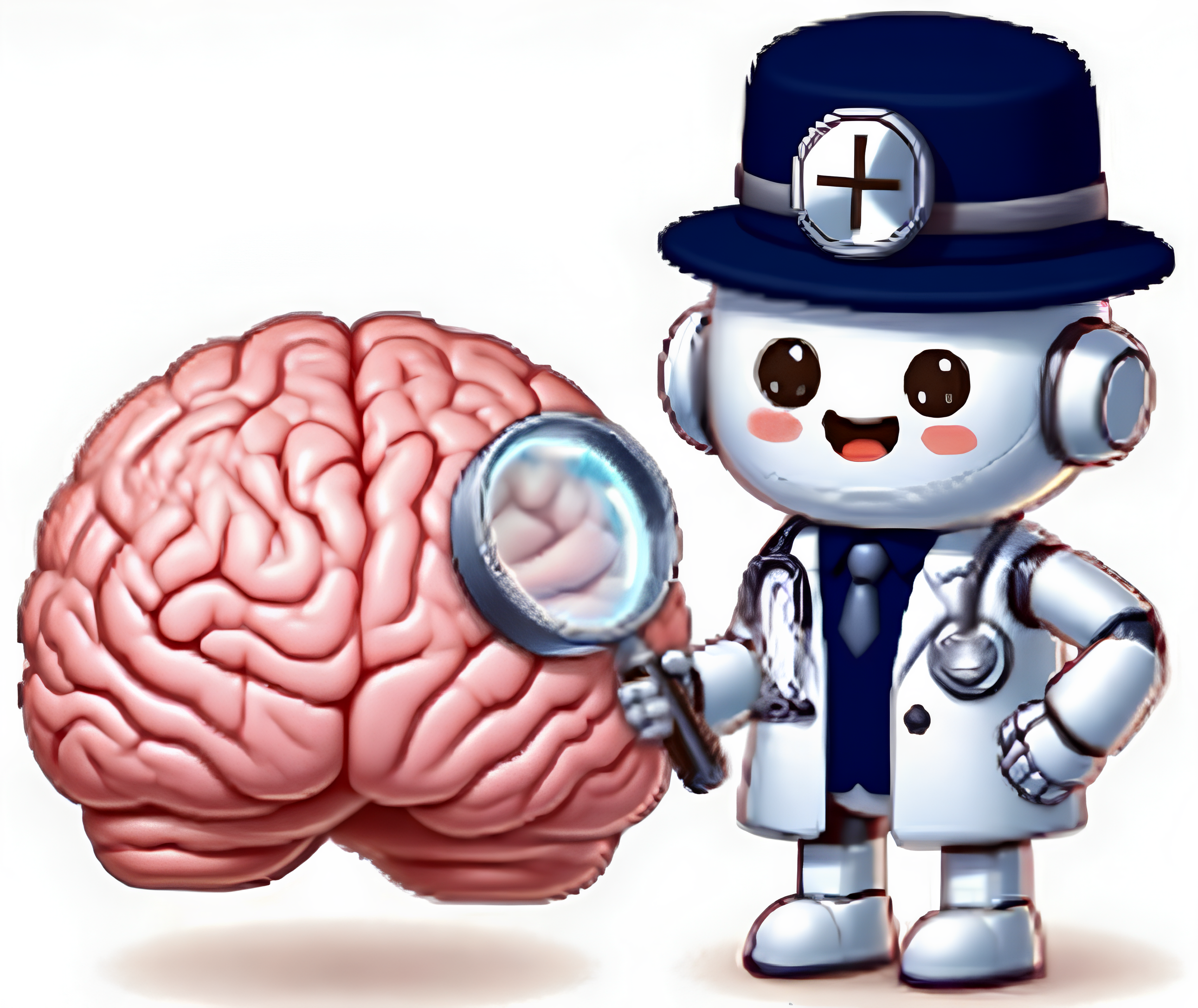}} OmniBrainBench: A Comprehensive Multimodal Benchmark for \\Brain Imaging Analysis Across Multi-stage Clinical Tasks}

\author{
Zhihao Peng$^{1*}$ \quad Cheng Wang$^{1*}$ \quad Shengyuan Liu$^{1*}$  \quad Zhiying Liang $^{2}\thanks{Equal contributions.}$ \quad Zanting Ye $^{3}$ \\ Minjie Ju $^{4}$ \quad  Peter YM Woo $^{5}$ \quad
Yixuan Yuan$^{1\thanks{Corresponding author (yxyuan@ee.cuhk.edu.hk)}}$ \\
 $^1$Chinese University of Hong Kong \quad
 $^2$Sun Yat-sen Memorial Hospital, Sun Yat-sen University \\
 $^3$Southern Medical University \quad
 $^4$ Zhongshan Hospital, Fudan University \\
 $^5$ Department of Neurosurgery, Prince of Wales Hospital \\
 \setcounter{footnote}{0}
}

\begin{document}
\maketitle
\begin{abstract}
{Brain imaging analysis is crucial for diagnosing and treating brain disorders, and multimodal large language models (MLLMs) are increasingly supporting it. However, current brain imaging visual question-answering (VQA) benchmarks either cover a limited number of imaging modalities or are restricted to coarse-grained pathological descriptions, hindering a comprehensive assessment of MLLMs across the full clinical continuum. To address these, we introduce~\ourbench, the first comprehensive multimodal VQA benchmark specifically designed to assess the multimodal comprehension capabilities of MLLMs in brain imaging analysis with closed- and open-ended evaluations.~\ourbench~comprises 15 distinct brain imaging modalities collected from 30 verified medical sources, yielding 9,527 validated VQA pairs and 31,706 images. It simulates clinical workflows and encompasses 15 multi-stage clinical tasks rigorously validated by a professional radiologist. Evaluations of 24 state-of-the-art models, including open-source general-purpose, medical, and proprietary MLLMs, highlight the substantial challenges posed by~\ourbench. Experiments reveal that proprietary MLLMs like GPT-5 (63.37\%) outperform others yet lag far behind physicians (91.35\%), while medical ones show wide variance in closed- and open-ended VQA. Open-source general-purpose MLLMs generally trail but excel in specific tasks, and all ones fall short in complex preoperative reasoning, revealing a critical visual-to-clinical gap.~\ourbench~establishes a new standard to assess MLLMs in brain imaging analysis, highlighting the gaps against physicians. We publicly release our benchmark at \href{https://cuhk-aim-group.github.io/OmniBrainBench.github.io/}{link}.
}

\end{abstract}

\section{Introduction}
\label{sec:intro}

{Brain imaging analysis has become a cornerstone of modern diagnostic and therapeutic decision-making by visualizing structural and functional abnormalities~\cite{abulnaga2025multimorph,shin2025anatomical,jack2024overview}, detecting early pathological changes~\cite{zhang2025incomplete,huang2025ai}, and supporting longitudinal monitoring of neurological diseases~\cite{lachinov2023learning,gutbrod2025openmibood}. In routine practice, traditional brain imaging analysis techniques heavily rely on the subjective expertise of physicians, causing variability and delays.} 
Recently, multimodal large language models (MLLMs)~\cite{guo2025deepseek,grok4_2025,openai2025gpt5,comanici2025gemini} have demonstrated attractive promise in multimodal perception, contextual understanding, and cross-modal reasoning with natural images, and are expected to significantly impact brain imaging analysis across diverse modalities. However, applying these models to brain imaging analysis presents domain-specific challenges. The scarcity of brain-specific expertise and the necessity to account for clinical-specific anatomical variations pose significant hurdles~\cite{liu2025comprehensive,johnson2019mimic,irvin2019chexpert}. A subsequent natural question is how to design a specialized brain imaging benchmark that aligns with multi-stage clinical workflows to evaluate the comprehension capability of MLLMs in brain imaging analysis~\cite{shehzad2025brain,bercea2025nova,tbkk-q937-25}.

{A major challenge in evaluating MLLMs is the limited modality coverage in existing brain imaging benchmarks~\cite{ray2024nova,seenivasan2022surgical,bai2024m3d,kucs2024medsegbench,wardlaw2019small}. Most benchmarks emphasize limited modalities and fail to fully cover the commonly used spectrum of structural, functional, and molecular neuroimaging~\cite{thomalla2018wake}. For example, Brain Tumor VQA~\cite{shehzad2025brain} is restricted to structural magnetic resonance imaging (sMRI) volumes, omitting functional series such as functional MRI (fMRI), diffusion imaging, and molecular techniques like positron emission tomography (PET). NOVA~\cite{ray2024nova} focuses on anatomical brain MRI and offers no coverage of nuclear medicine or other modalities. In practice, clinical reality instead demands modality diversity~\cite{mahler2020multimodality,tang2024multimodal}, e.g., stroke evaluation begins with non-contrast computed tomography (CT) to rule out hemorrhage and adds diffusion-weighted imaging (DWI), susceptibility-weighted imaging (SWI), and fluid-attenuated inversion recovery (FLAIR) to map the damaged tissue and adjacent swelling~\cite{wardlaw2019small}. Parkinson's disease management integrates sMRI, diffusion MRI, resting-state fMRI, and dopaminergic imaging via single-photon emission computed tomography (SPECT) or PET to track brain degeneration and functional network changes~\cite{elzahraa2023multimodality}. Alzheimer's disease assessment relies on T1-weighted (T1W) imaging with PET to capture structural atrophy and protein deposition~\cite{liu2025plasma}. There is a clear gap between existing benchmarks and clinical practice in neuroimaging, underscoring the need for a comprehensive evaluation framework that supports multimodal assessment.
}



{
Another challenge is that existing brain imaging benchmarks focus on limited tasks. Factually, a complete clinical workflow begins with anatomical identification, advances to lesion localization, and then proceeds to treatment planning and prognostic assessment. Yet current benchmarks cover only a subset of these stages, e.g., Brain Tumor VQA~\cite{shehzad2025brain} is limited to tumor type and basic attributes, omitting prognosis estimation and treatment planning. VQA-RAD~\cite{lau2018dataset} emphasizes basic findings without integrated evaluation across localization, diagnosis, and prognosis. NOVA~\cite{ray2024nova} concentrates on lesion localization in MRI without linking results to outcome prediction and management. In clinical practice, end-to-end competence is necessary, e.g., assessment in high-grade gliomas is standardized by criteria that consider tumor components, clinical status, and medication use, so that radiographic changes can be interpreted reliably for trial endpoints and patient management~\cite{wen2010rano}. Presurgical evaluation of drug-resistant epilepsy leverages multimodal imaging and analytic algorithms to identify brain regions where seizures originate, assess surgical risk, and plan interventions~\cite{eanm2024_epilepsy}. These examples show that evaluating MLLMs on only a subset of stages underestimates the broad skills needed for comprehensive brain imaging analysis. Therefore, a benchmark that aligns with the full range of clinical needs and can verify how models will perform across real-world tasks is highly demanded.
}

\begin{figure*}[htbp]
    \centering
    
    \begin{subfigure}[t]{0.92\linewidth}
        \centering
        \includegraphics[width=\linewidth]{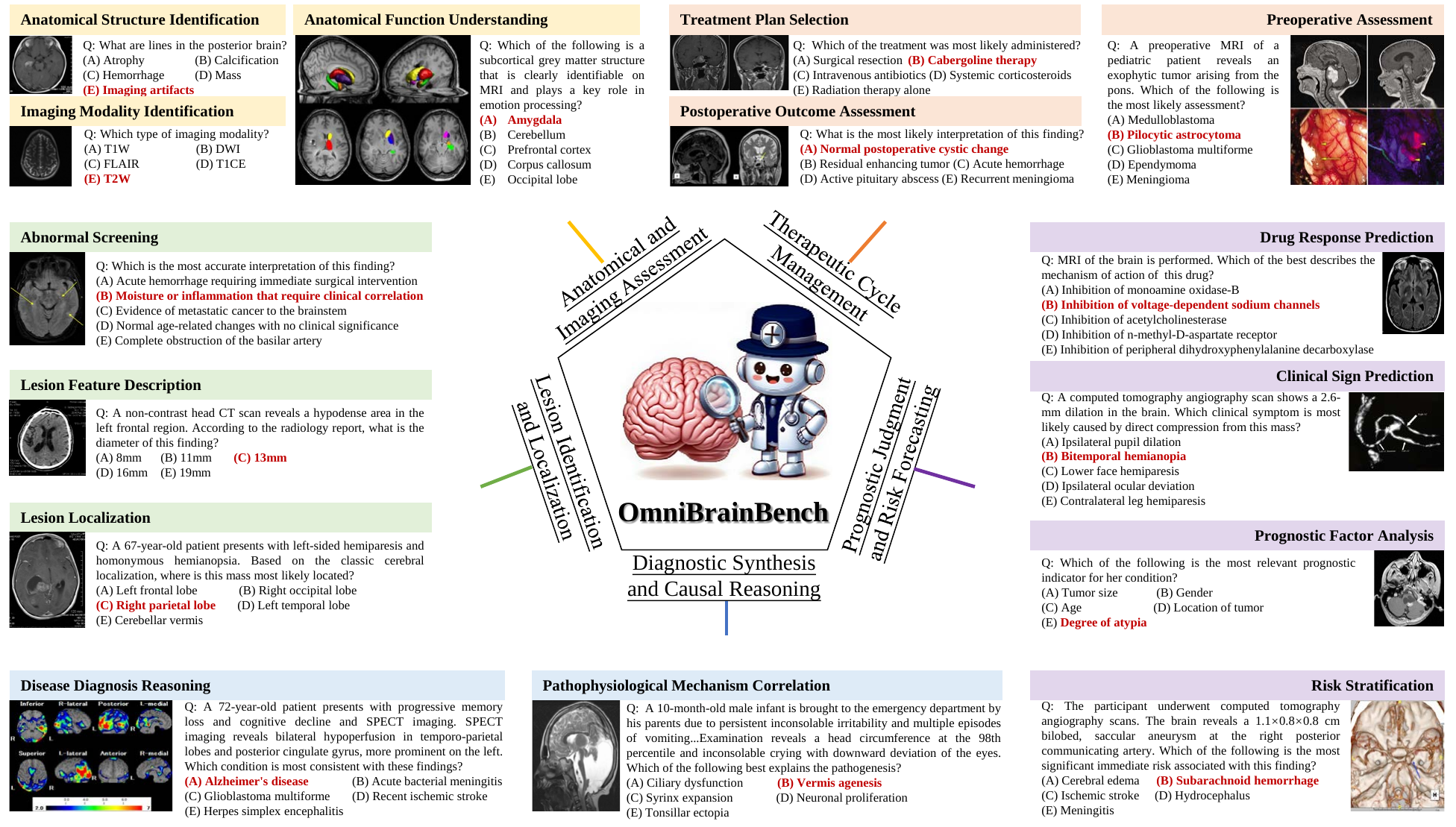}
        \includegraphics[width=\linewidth]{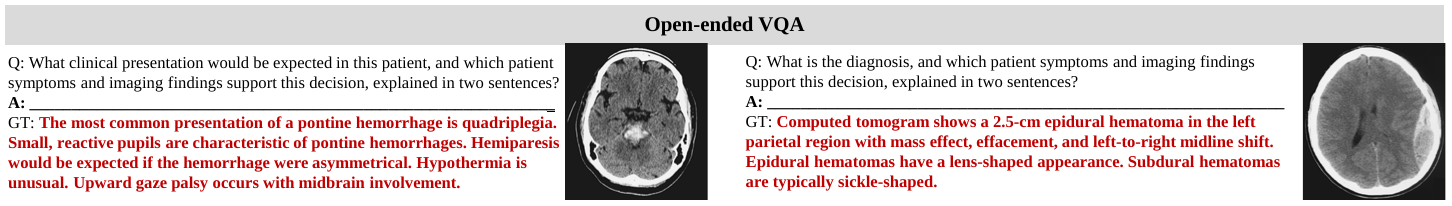}
        \caption{\textbf{Overview}}
        \label{fig:1a}
    \end{subfigure}
    \hfill
    \begin{subfigure}[t]{0.44\linewidth}
        \centering
        \includegraphics[width=\linewidth]{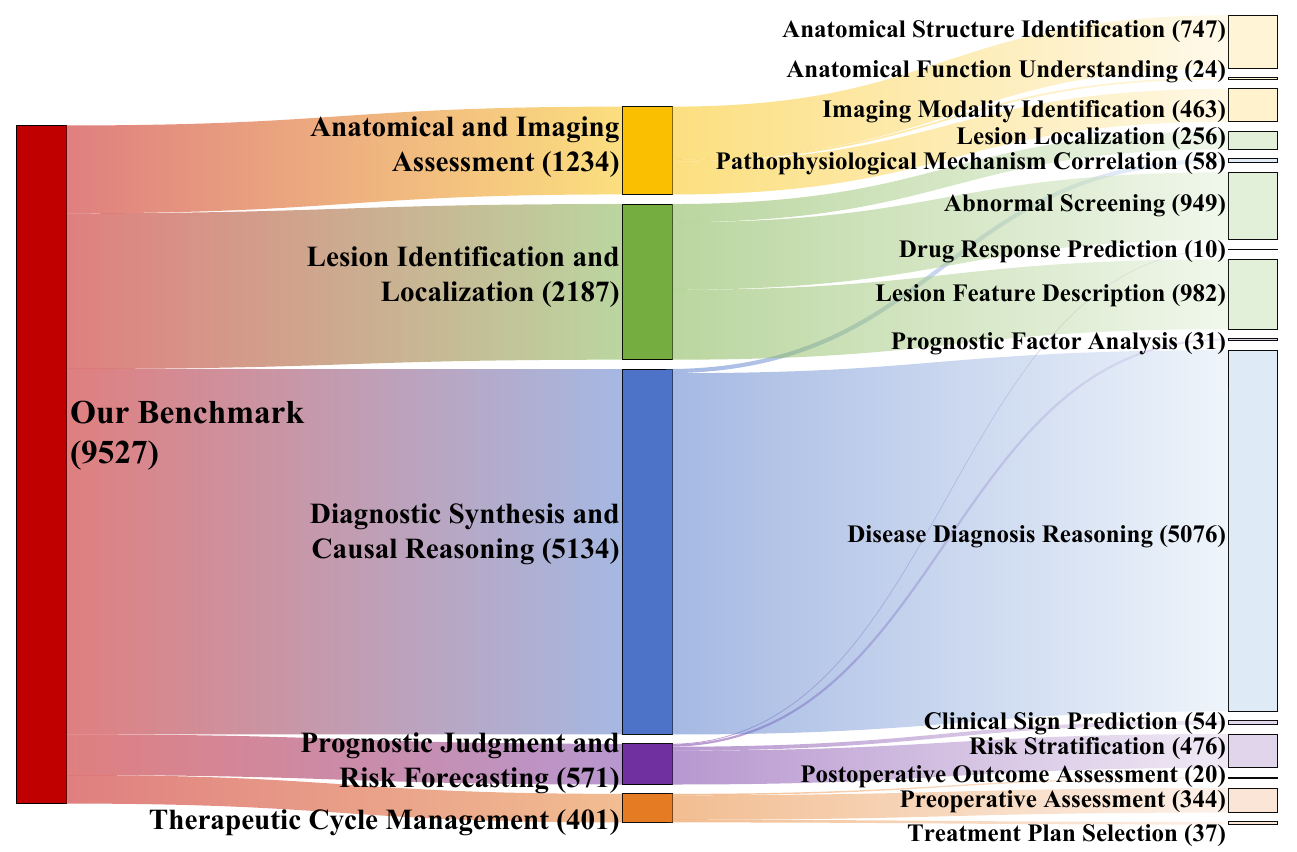}
        \caption{\textbf{Diverse Tasks Distributions}}
        \label{fig:1b}
    \end{subfigure}
    \hfill
    \begin{subfigure}[t]{0.50\textwidth}
        \centering
        \includegraphics[width=\linewidth]{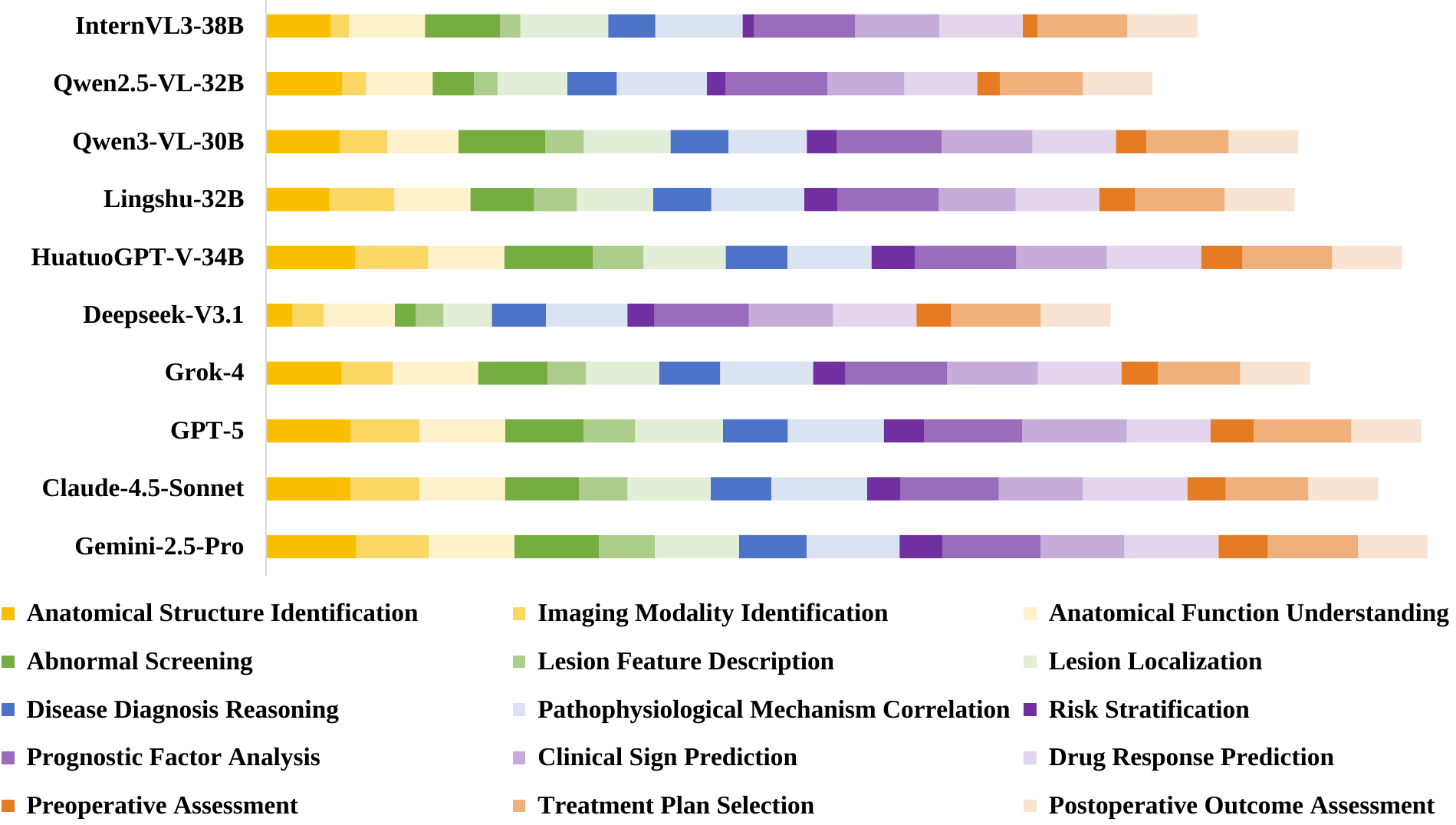}
        \caption{\textbf{Multi-dimensional Evaluation}}
        \label{fig:1c}
    \end{subfigure}
    \caption{Overview of our~\ourbench, the first comprehensive multimodal benchmark specifically designed to evaluate MLLMs across the complete spectrum of brain imaging analysis, covering 15 distinct clinical scenarios, drawn from 30 rigorously validated sources, comprising 9,527 clinically verified VQA pairs and 31,706 images.}
    \label{fig:example}
\end{figure*}

To address these challenges, we introduce~\textbf{\ourbench}, the comprehensive multimodal benchmark for evaluating MLLMs in brain imaging analysis, as shown in Fig.~\ref{fig:example}. To the best of our knowledge,~\ourbench~is the most extensive multimodal brain imaging benchmark to date, drawn from 30 rigorously validated sources, yielding 259,628 instruction-tuning collection comprising our~\ourvqa, the current largest brain imaging instruction-tuning collection. From~\ourvqa, we extract representative pairs that undergo rigorous clinical validation by a radiologist with over 13 years of experience, resulting in our final~\ourbench~of 9,527 clinically verified VQA pairs and 31,706 images, as detailed in Fig.~\ref {fig: data}. To further distinguish the difference between~\ourbench~and other existing ones, we elaborate the benchmark details as detailed in Table \ref{tab:comp_other_bench}. For \textbf{\textit{modality coverage}}, the benchmark provides 15 imaging modalities, including the coarse-grained ones include CT, MRI, PET, SPECT, anatomical diagram (ADiag), histopathology imaging (HI) modalities, and the fine-grained ones include DWI, SWI, FLAIR, T1W, T1-weighted contrast-enhanced (T1CE), T2-weighted imaging (T2W), magnetic resonance angiography (MRA), proton density weighted imaging (PD), fMRI. With its extensive scope and multi-dimensional evaluation criteria,~\ourbench~is positioned to comprehensively assess the effectiveness of MLLMs in diverse brain imaging modality data. For \textbf{\textit{clinical tasks}}, Our~\ourbench~is designed to assess the multimodal comprehension capabilities of MLLMs across the full clinical continuum, spanning five specialized clinical phases (i.e., anatomical structure identification, disease diagnosis reasoning, lesion localization, prognostic factor analysis, and postoperative outcome assessment) within 15 multi-stage clinical tasks, as detailed in Fig.~\ref{fig:example}. These tasks are rigorously validated by several physicians, and it is expected to span from basic anatomical recognition to complex diagnostic synthesis, prognostic judgment, and therapeutic cycle management. 

We benchmark 24 state-of-the-art models, including open-source general-purpose, medical-specialized, and proprietary MLLMs, with the closed-ended evaluation comprising 6,823 multiple-choice VQA pairs with five options and one correct answer, and the open-ended evaluation consisting of 2,704 free-form descriptive VQA pairs that require detailed clinical reasoning. By using human clinician performance as a reference, the comparisons highlight gaps in perceiving, understanding, and reasoning between MLLMs and physicians in brain imaging analysis. The main contributions are summarized as follows:
\begin{itemize}
    \item We introduce~\textbf{\ourbench}, the first comprehensive multimodal benchmark specifically designed to evaluate MLLMs across the complete spectrum of brain imaging analysis with closed- and open-ended evaluations, covering {\textbf{9,527} clinically verified VQA pairs, \textbf{31,706} images, and \textbf{15} modalities}.
    \item We develop a multi-dimensional evaluation framework that mirrors the clinical workflow from anatomical and imaging assessment to therapeutic cycle management, assessing the capabilities of MLLMs across \textbf{15 multi-stage clinical tasks} within brain imaging analysis.
    \item We conduct {\textbf{extensive evaluations of 24 models}} across open-source general-purpose, medical-specialized, and proprietary MLLMs to reveal critical gaps in their visual-clinical reasoning, providing a detailed analysis of MLLMs in brain imaging.
\end{itemize}

\begin{table*}[]
\centering
\caption{Comparisons with existing multimodal brain imaging benchmarks. AIA indicates Anatomical and Imaging Assessment; LIL indicates Lesion Identification and Localization; DSCR indicates Diagnostic Synthesis and Causal Reasoning; PJRF indicates Prognostic Judgment and Risk Forecasting; TCM indicates Therapeutic Cycle Management; The asterisk ($*$) denotes the brain imaging data of the benchmark, where we mark it to distinguish neuroimaging samples from non-brain medical samples (e.g., chest X-ray, abdominal CT, etc).}
\resizebox{0.88\textwidth}{!}{
\begin{tabular}{l|ccccccccccc}
\hline\hline
\textbf{Benchmark}                      & \textbf{\begin{tabular}[c]{@{}l@{}}Closed\\ -ended\end{tabular}} & \textbf{\begin{tabular}[c]{@{}l@{}}Open\\ -ended\end{tabular}} & \textbf{Images} & \textbf{QA pairs} & \textbf{Modality} & \textbf{Task} & \textbf{AIA}                  & \textbf{LIL}                  & \textbf{DSCR}                 & \textbf{PJRF}                 & \textbf{TCM}                  \\\hline
MMMU-Pro* \cite{yue-etal-2025-mmmu}     &            \textcolor{black}{\Checkmark}                         &                                                                & 18              & 18                & 1                 & 3             & \textcolor{black}{\Checkmark} & \textcolor{black}{\Checkmark} & \textcolor{black}{\Checkmark} &                               &                               \\
MedXpertQA* \cite{zuo2025medxpertqa}    &            \textcolor{black}{\Checkmark}                         &                                                                & 42              & 34                & 6                 & 8             & \textcolor{black}{\Checkmark} & \textcolor{black}{\Checkmark} & \textcolor{black}{\Checkmark} &                               & \textcolor{black}{\Checkmark} \\
NEJMIC* \cite{nejm2025imagechallenge}   &            \textcolor{black}{\Checkmark}                         &                                                                & 42              & 35                & 3                 & 5             &                               &                               & \textcolor{black}{\Checkmark} & \textcolor{black}{\Checkmark} & \textcolor{black}{\Checkmark} \\
VQA-RAD* \cite{lau2018dataset}          &            \textcolor{black}{\Checkmark}                         &                   \textcolor{black}{\Checkmark}                & 55              & 99                & 2                 & 6             & \textcolor{black}{\Checkmark} & \textcolor{black}{\Checkmark} & \textcolor{black}{\Checkmark} &                               &                               \\
Slake* \cite{liu2021slake}              &            \textcolor{black}{\Checkmark}                         &                   \textcolor{black}{\Checkmark}                & 165             & 3,148             & 3                 & 7             & \textcolor{black}{\Checkmark} & \textcolor{black}{\Checkmark} &                               &                               & \textcolor{black}{\Checkmark} \\
MMMU* \cite{yue2024mmmu}                &            \textcolor{black}{\Checkmark}                         &                   \textcolor{black}{\Checkmark}                & 337             & 309               & 4                 & 5             & \textcolor{black}{\Checkmark} & \textcolor{black}{\Checkmark} & \textcolor{black}{\Checkmark} &                               &                               \\
Br35H \cite{tbkk-q937-25}               &            \textcolor{black}{\Checkmark}                         &                                                                & 339             & 339               & 1                 & 2             & \textcolor{black}{\Checkmark} &                               & \textcolor{black}{\Checkmark} &                               &                               \\
Brain Tumor VQA \cite{shehzad2025brain} &            \textcolor{black}{\Checkmark}                         &                   \textcolor{black}{\Checkmark}                & 750             & 14,015            & 1                 & 3             &                               & \textcolor{black}{\Checkmark} & \textcolor{black}{\Checkmark} &                               &                               \\
MedFrameQA* \cite{yu2025medframeqa}     &            \textcolor{black}{\Checkmark}                         &                                                                & 823             & 292               & 3                 & 6             &                               & \textcolor{black}{\Checkmark} & \textcolor{black}{\Checkmark} &                               & \textcolor{black}{\Checkmark} \\
NOVA \cite{bercea2025nova}              &                                                                  &                   \textcolor{black}{\Checkmark}                & 906             & 281               & 1                 & 4             &                               &                               & \textcolor{black}{\Checkmark} & \textcolor{black}{\Checkmark} & \textcolor{black}{\Checkmark} \\
RadImageNet* \cite{mei2022radimagenet}  &            \textcolor{black}{\Checkmark}                         &                                                                & 2,037           & 2,079             & 1                 & 2             & \textcolor{black}{\Checkmark} &                               & \textcolor{black}{\Checkmark} &                               &                               \\
OmniMedVQA* \cite{hu2024omnimedvqa}     &            \textcolor{black}{\Checkmark}                         &                                                                & 2,376           & 2,418             & 1                 & 2             & \textcolor{black}{\Checkmark} &                               & \textcolor{black}{\Checkmark} &                               &                               \\
PMC-VQA* \cite{zhang2023pmc}            &            \textcolor{black}{\Checkmark}                         &                   \textcolor{black}{\Checkmark}                & 10,799          & 12,591            & 12                & 7             & \textcolor{black}{\Checkmark} & \textcolor{black}{\Checkmark} & \textcolor{black}{\Checkmark} &                               & \textcolor{black}{\Checkmark} \\
PubMedVision* \cite{chen2024huatuogpt}  &                                                                  &                   \textcolor{black}{\Checkmark}                & 34,929          & 53,554            & 3                 & 2             & \textcolor{black}{\Checkmark} & \textcolor{black}{\Checkmark} & \textcolor{black}{\Checkmark} &                               &                               \\ \hline
\ourvqa                                 & \multicolumn{1}{c}{\textcolor{black}{\Checkmark}}                & \multicolumn{1}{c}{\textcolor{black}{\Checkmark}}              & 600,050         & 259,628           & 15                & 15            & \textcolor{black}{\Checkmark} & \textcolor{black}{\Checkmark} & \textcolor{black}{\Checkmark} & \textcolor{black}{\Checkmark} & \textcolor{black}{\Checkmark} \\
\ourbench                               & \multicolumn{1}{c}{\textcolor{black}{\Checkmark}}                & \multicolumn{1}{c}{\textcolor{black}{\Checkmark}}              & 31,706          & 9,527             & 15                & 15            & \textcolor{black}{\Checkmark} & \textcolor{black}{\Checkmark} & \textcolor{black}{\Checkmark} & \textcolor{black}{\Checkmark} & \textcolor{black}{\Checkmark} \\
\hline\hline
\end{tabular}
}
\label{tab:comp_other_bench}
\end{table*}

\section{Related Work}
\label{sec:related_work}

\subsection{MLLMs}
{
Early MLLMs (e.g., CLIP~\cite{radford2021clip}, BLIP~\cite{li2022blip,li2023blip2}, Flamingo~\cite{alayrac2022flamingo}) fuse a frozen language backbone with a vision encoder and lightweight projector to align visual features to text, enabling VQA with few-shot generalization. Recent general MLLMs like Janus-Pro~\cite{chen2025janus}, InternVL3~\cite{zhu2025internvl3}, QwenVL~\cite{qwen2.5-VL}, Deepseek-V3~\cite{deepseekai2024deepseekv3technicalreport,guo2025deepseek}, Grok-4~\cite{grok4_2025}, GPTs~\cite{openai2025gpt5}, Claude-4.5-Sonnet~\cite{anthropic2025claude45} and Gemini-2.5-Pro~\cite{comanici2025gemini} jointly train on web-scale interleaved data using contrastive and generative objectives, yielding strong CoT reasoning, tool use, and long-context handling across images, documents, and video. The medical field quickly adapts these~\cite{ye2025multimodal}: MedVLM-R1~\cite{pan2025medvlm} adds radiology supervision, LLaVA-Med~\cite{li2023llava} injects biomedical dialogues, Lingshu~\cite{xu2025lingshu} tunes bilingually, and HuatuoGPT~\cite{chen2024huatuogpt} scales to millions of image-text pairs to curb hallucination and boost diagnosis. Overall, MLLM evolution shifts from cross-modal alignment to domain instruction tuning and retrieval-augmented generation, from 2D to 3D analysis, and from perception to end-to-end clinical reasoning, prioritizing safety, factuality, and rigorous evaluation.
}

\noindent \subsection{Benchmark for Medical MLLMs}
Benchmarks have mirrored this evolution from small-scale, single-modality QA to comprehensive multi-modality, multi-task evaluation~\cite{yue2024mmmu,yue-etal-2025-mmmu,shehzad2025brain,yu2025medframeqa,hu2024omnimedvqa,zhang2023pmc}. For instance, VQA-RAD~\cite{lau2018dataset} targeted basic clinical QA on 2D radiology images, while MIMIC-CXR~\cite{johnson2019mimic} and CheXpert~\cite{irvin2019chexpert} enabled report generation and broader QA but stayed chest- and 2D-centric. Recent efforts have broadened the modality and task scope, with MedTrinity-25M~\cite{xie2024medtrinity} offering multimodal, multi-granularity supervision for scaled instruction tuning. Yet, these were not tailored for neuroimaging; brain scans form only a minor portion of the data. New brain-specific benchmarks have emerged, e.g., NOVA~\cite{ray2024nova} emphasizes anomaly localization and OOD reasoning, but it risks bias from single-source data. Segmentation suites like MedSegBench~\cite{kucs2024medsegbench} or BraTS~\cite{menze2015multimodal} focus on perception, not end-to-end neuro-clinical reasoning. Overall, medical MLLM benchmarks have advanced from few-modality, single-task settings to large-scale, multimodal, multi-task settings with 3D volumes; yet, brain-oriented ones still lack full modality coverage and clinical alignment.

\section{\ourbench~}
\subsection{Overview}

{
\ourbench~is a comprehensive benchmark that comprises 9,527 clinically validated VQA pairs with 31706 images to assess the perception, understanding, and reasoning skills of MLLMs across a broad scope of clinical scenarios, detailed in Table \ref{tab:comp_other_bench}. It covers 15 specialized tasks across five clinical phases that reflect the progression of diagnostic and therapeutic decision-making within brain imaging analysis, meticulously designed to align with the following clinical process: ``anatomical and imaging assessment\textbf{$\rightarrow$}lesion identification and localization\textbf{$\rightarrow$}diagnostic synthesis and causal reasoning\textbf{$\rightarrow$}prognostic judgment and risk forecasting\textbf{$\rightarrow$}therapeutic cycle management''. Each phase includes specialized tasks and is closely interconnected. This structural framework enables us to precisely evaluate how well MLLMs can perceive, understand, and reason information across diverse brain imaging data to derive the solution.
}

\begin{figure}[t]
  \centering
   \includegraphics[width=0.98\linewidth]{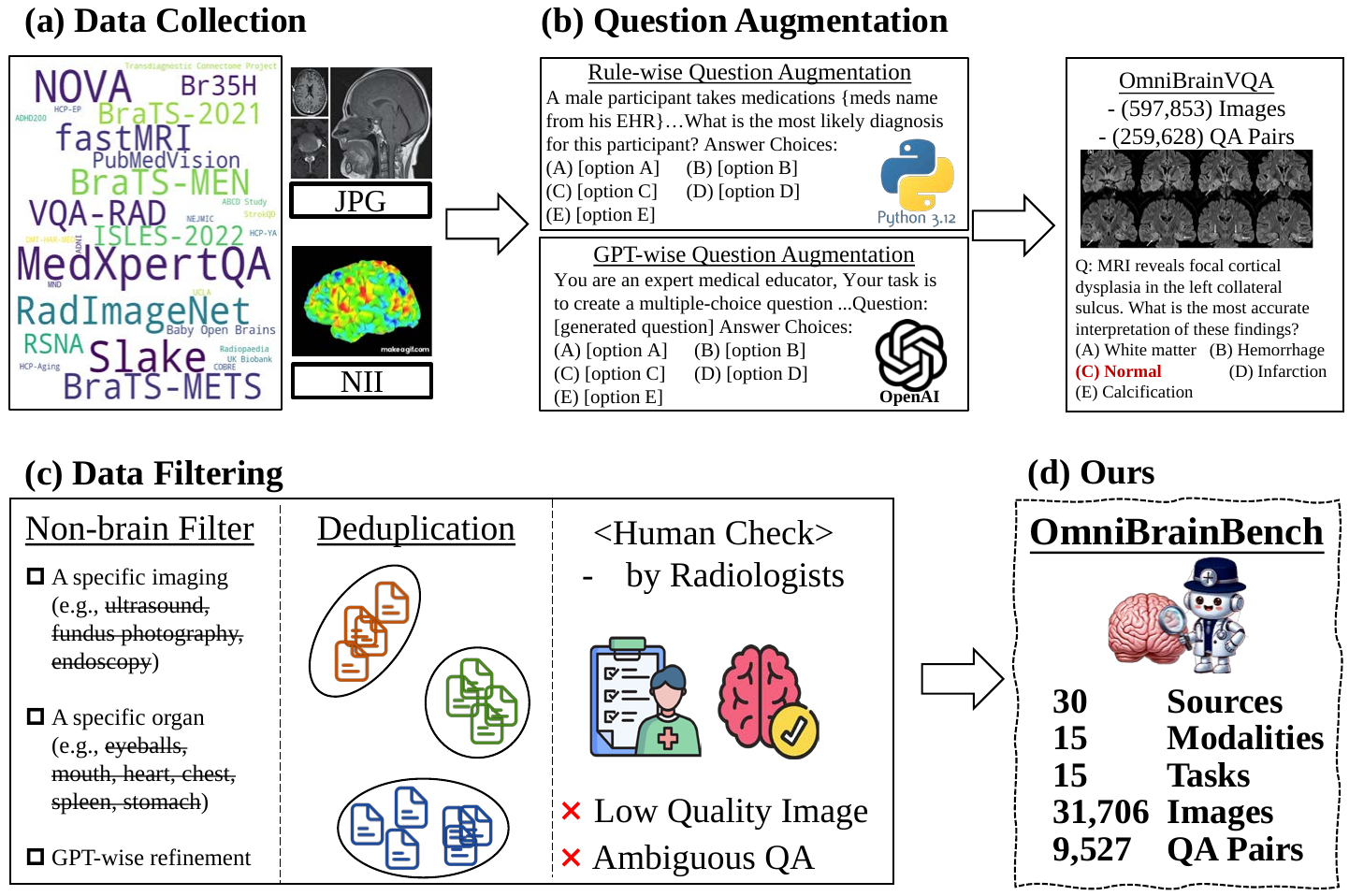}
   \caption{Construction process of~\textbf{\ourbench} with (a) data collection, (b) question augmentation, and (c) data filtering.}
   \label{fig: data}
\end{figure}

\subsection{Construction Pipeline}
{In this part, we introduce the benchmark construction process of~\ourbench, as shown in Fig.~\ref{fig: data}.
}

\begin{figure}[t]
  \centering
   \includegraphics[width=0.92\linewidth]{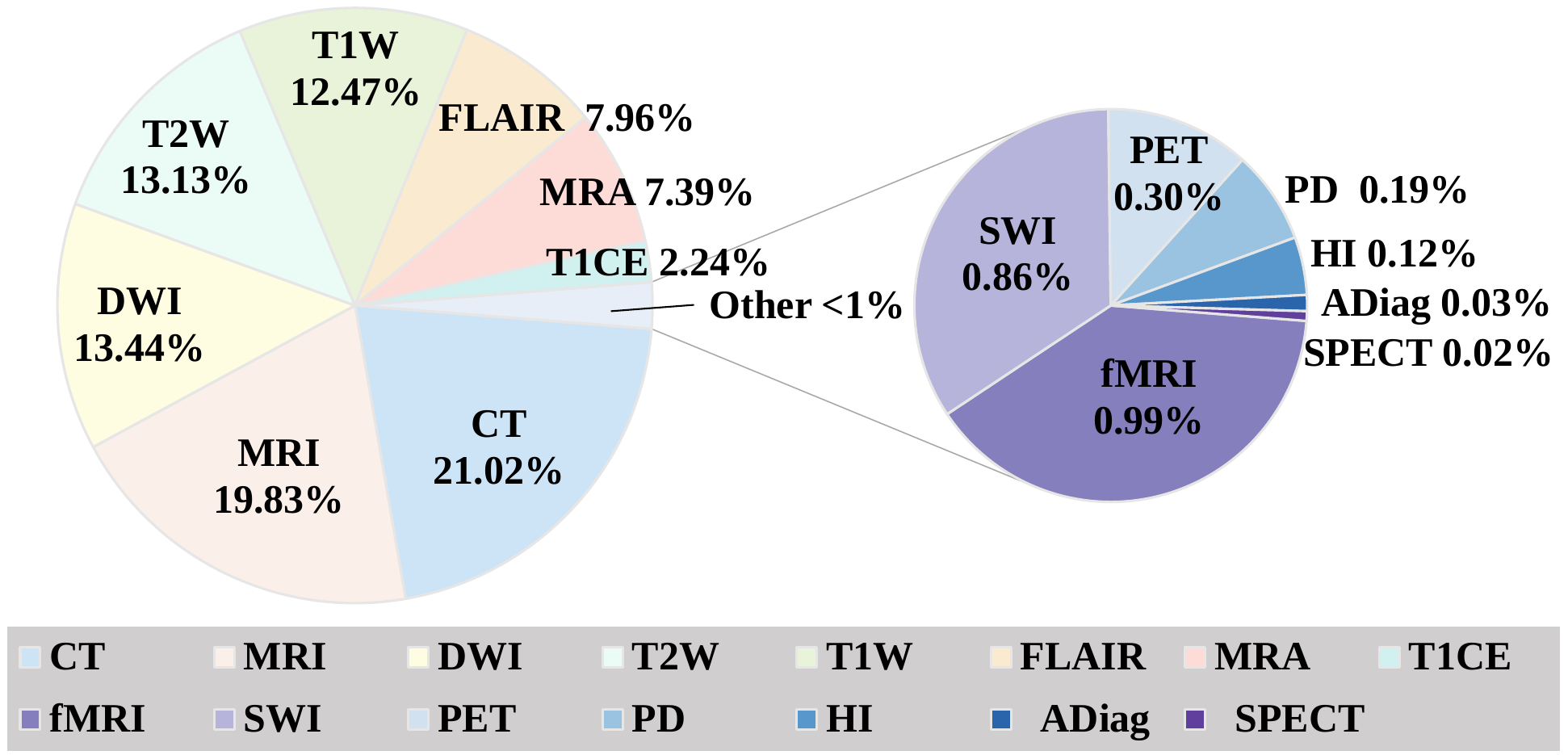}
   \caption{Modality Distribution.}
   \label{fig: Modality Distribution}
\end{figure}

\noindent\textbf{Data Collection.} To ensure comprehensive coverage, we gather 30 public brain imaging datasets from online sources (DICOM, JSON, XLS, CSV, JPG, PNG, NII) to encompass various brain imaging types and terminologies, including BraTS-MEN \cite{labella2023asnr}, BraTS-METS \cite{moawad2024brain}, fastMRI \cite{zbontar2018fastmri}, VQA-RAD \cite{lau2018dataset}, BraTS-2021 \cite{baid2021rsna}, ISLES-2022 \cite{hernandez2022isles}, Br35H \cite{tbkk-q937-25}, RadImageNet \cite{mei2022radimagenet}, RSNA \cite{stein2019rsna}, NOVA \cite{bercea2025nova}, PubMedVision \cite{chen2024huatuogpt}, Baby Open Brains \cite{feczko2025baby}, TCP \cite{chopra2025transdiagnostic}, NEJMIC \cite{nejm2025imagechallenge}, MedXpertQA \cite{zuo2025medxpertqa}, Slake \cite{liu2021slake}, MND~\cite{chang2025fmri}, ABCD~\cite{casey2018adolescent}, ADHD200~\cite{adhd2012adhd}, COBRE~\cite{calhoun2012exploring}, DMT-HAR-MED~\cite{meling2024meditating}, HCP-Aging~\cite{van2013wu}, HCP-EP~\cite{van2013wu}, HCP-YA~\cite{van2013wu}, UCLA~\cite{poldrack2016phenome}, ADHD200~\cite{brown2012adhd}, UK Biobank~\cite{palmer2007uk}, ADNI~\cite{mueller2005alzheimer}, Radiopaedia~\cite{radiopaedia}, and StrokQD~\cite{zhang2023automatic}. It incorporates 15 modality labels (Fig.~\ref{fig: Modality Distribution}), spanning coarse- and fine-grained professional terminology, where the coarse-grained ones include CT, MRI, PET, SPECT, ADiag, HI modalities, and the fine-grained ones include T2W, FLAIR, DWI, SWI, T1W, T1CE, MRA, PD, fMRI. There is a hierarchical relationship between them, where coarse-grained terms are parent categories, and fine-grained terms are child categories. For example, MRI (coarse-grained) includes T1W, T2W, FLAIR, DWI, fMRI (fine-grained), etc. For raw 3D data, we consulted board-certified radiologists and adopted a commonly used strategy to select 2D slice images from 3D volume data along axial, sagittal, and coronal anatomical planes. Notably, we select NEJMIC, Radiopaedia, and StrokQD as sources for the open-ended VQA because they contain clinical reasoning information provided by expert clinicians. We assemble them into the structured clinical format to evaluate MLLMs capability.

\noindent\textbf{Question Augmentation.} {Following data collection, we implement a systematic question augmentation to integrate the metadata of datasets, involving rule-based and GPT-based approaches. Specifically, for disease-specific and modality-specific cases, we extract metadata from their clinical documentation and generate questions and options using a standardized template, as detailed in Fig.~\ref{fig: data} (b). For cases with multi-granular textual descriptions, we utilize a flexible GPT-based approach via the GPT-5 API to create plausible distractors, resulting in a multiple-choice format with five options per question, ensuring all options are plausible for medical professionals. Prompt engineering, outlined in Fig.~\ref{fig: data} (b), enforces consistent requirements and output formats. This process yields 259,628 clinically VQA pairs with 600,050 images, forming our~\ourvqa.}

\noindent\textbf{Data Filtering.} To create a representative and balanced subset for MLLM evaluation, we implement a systematic data filtering pipeline on our~\ourvqa~dataset to obtain our~\ourbench. We first filter JSON records containing non-brain content from specific imaging modalities (e.g., ultrasound, fundus photography, optical coherence tomography, endoscopy, dermoscopy) and non-brain systems (e.g., breast, cardiac, chest, gastrointestinal, hematology, hepatobiliary, musculoskeletal, spine, urogenital, vascular). Next, we apply GPT-based refinement using the GPT-5 API to reformulate questions from the original QA pairs, varying expression styles while preserving semantic content, enhancing the adaptability assessment of MLLMs to diverse style representations. Subsequently, we encode texts with Sentence Transformers~\cite{reimers2019sentence} and images with DINO-V2~\cite{oquab2024dinov2} to extract textual and visual embeddings for data deduplication. From each group, we select the question-answer pair closest to the centroid, ensuring a representative and diverse sample for evaluation. 

To reflect the complexity of clinical scenarios and skills required in clinical workflows, we developed 15 specialized clinical tasks across five core phases of clinical workflows, validated through rigorous clinical review with a radiologist with over 13 years of experience. Each QA pair was mapped to its most relevant task using a curated prompt engineering template, as detailed in the Appendix. Furthermore, a subset of QA pairs incorporated multiple images to enhance the capability validation of MLLMs to handle complex clinical reasoning. These processes yielded our~\ourbench, comprising 9,527 VQA pairs in both closed- and open-ended formats, where the closed-ended evaluation comprises 6,823 multiple-choice VQA pairs with five options and one correct answer, and the open-ended evaluation consists of 2,704 descriptive VQA pairs based on structured clinical reports provided by expert clinicians.

\subsection{Clinical Modality Coverage}
The foundation of our benchmark is based on a comprehensive collection of imaging data, systematically categorized into five primary groups according to their clinical utility. The first group includes foundational structural images (e.g., CT, MRI, T1W, T2W, PD, and FLAIR), which establish anatomical baselines and detect gross abnormalities. The second group comprises pathology-sensitive structural images (e.g., FLAIR, DWI, SWI, MRA, and T1CE) for precise detection and characterization. The third group encompasses functional and molecular imaging modalities (e.g., PET, fMRI, and SPECT) to elucidate the etiology of the disease and the underlying pathophysiological mechanisms. The fourth group includes connectivity and metabolic imaging (e.g., PET, fMRI, SPECT, and DWI) to predict disease progression and associated risks. The fifth group consists of multimodal and serial imaging (e.g., ADiag and HI images) to guide treatment planning, procedural interventions, outcome monitoring, etc.


\begin{table*}[]
\centering
\caption{Performance of different MLLMs on five specialized clinical phases with 15 secondary subtasks on closed-ended VQA of~\ourbench~. The best-performing model in each category is highlighted in \textbf{bold}, and the second best is highlighted in \underline{underlined}. The definition of the abbreviation is provided in Sec. \ref{subsec:multitasks}.}
\resizebox{0.92\linewidth}{!}{
\begin{tabular}{l|ccc|ccc|cc|cccc|ccc|c}
\hline\hline
\multirow{2}{*}{MLLMs} & \multicolumn{3}{c|}{AIA} & \multicolumn{3}{c|}{LIL} & \multicolumn{2}{c|}{DSCR} & \multicolumn{4}{c|}{PJRF} & \multicolumn{3}{c|}{TCM} & \multirow{2}{*}{Overall} \\ \cline{2-16}
                       & ASI    & IMI    & AFU   & AS     & LFD    & LL    & DDR         & PMC        & RS  & PFA  & CSP  & DRP  & PA    & TPS    & POA    &                          \\
\hline
Physician                 &  100.00               & 100.00                & 100.00                & 100.00                & 80.00                & 100.00                &   89.71              & 90.00               & 80.00                &  90.00                & 90.00                 & 90.00                 & 80.00                &  100.00               & 90.00                 &   91.35                       \\
\rowcolor{OrangeRed!20} 
\multicolumn{17}{c}{\textbf{Open-Source General-Purpose MLLMs}} \\ 
Janus-Pro-7B              & 61.85          & 49.68          & 63.64          & 68.19          & 36.23          & 73.12          & 38.95          & 66.67          & 25.47          & 90.32           & 80.00           & 70.00           & 24.44          & 78.57          &\underline{83.33}           & 45.11                   \\
InternVL3-8B              & 70.55          & 46.65          & 72.73          & \textbf{91.46} & 37.68          & 79.57          & 47.49          & 80.56          & 21.26          & 87.10           & 80.00           & 70.00           & 31.43          & 71.43          & 66.67           & 53.25                   \\
InternVL3-9B              & 61.31          & 20.95          & 72.73          & 58.17          & 35.14          & 75.27          & 55.30          & 75.00          & 33.05          & 93.55           & 80.00           & 80.00           & 37.46          & 71.43          & 66.67           & 51.52                   \\
InternVL3-14B             & 56.76          & 16.20          & 68.18          & 86.14          & 31.88          & \textbf{86.02} & 53.16          & 83.33          & 28.42          & 90.32           & 86.67           & \underline{90.00}           & 31.75          & 64.29          & 66.67           & 52.37                   \\
InternVL3-38B             & 61.58          & 17.49          & 72.73          & 71.29          & 19.57          & \underline{83.87}          & 44.79          & 83.33          & 10.53          & \underline{96.77}           & 80.00           & 80.00           & 13.97          & \underline{85.71}          & 66.67           & 44.38                   \\
Qwen2.5-VL-7B             & 74.83          & 37.37          & 72.73          & 62.87          & 33.33          & 68.82          & 45.87          & 80.56          & 24.63          & 90.32           & 86.67           & 70.00           & 30.48          & 64.29          & 66.67           & 48.75                   \\
Qwen2.5-VL-32B            & 72.69          & 22.68          & 63.64          & 39.48          & 22.46          & 66.67          & 46.87          & 86.11          & 17.89          & \underline{96.77}           & 73.33           & 70.00           & 21.59          & 78.57          & 66.67           & 43.94                   \\
Qwen3-VL-4B               & 76.04          & 32.83          & \underline{77.27}         & 63.61          & 33.33          & 69.89          & 48.97          & 66.67          & 26.11          & \underline{96.77}           & 80.00           & \underline{90.00}           & 32.06          & 71.43          & 66.67           & 50.45                   \\
Qwen3-VL-8B               & 75.50          & 38.01          & 68.18          & 62.38          & 28.99          & 76.34          & 47.92          & 77.78          & 16.00          & \underline{96.77}           & \underline{93.33}           & \underline{90.00}           & 22.22          & \underline{85.71}          & 66.67           & 48.89                   \\
Qwen3-VL-30B              & 70.15          & 45.36          & 68.18          & 82.80          & 36.59          & 82.80          & 55.07          & 75.00          & 28.21          & \textbf{100.00} & 86.67           & 80.00           & 28.57          & 78.57          & 66.67           & 56.40                   \\ 
\rowcolor{magenta!20} 
\multicolumn{17}{c}{\textbf{Open-Source Medical MLLMs}} \\ 
MedVLM-R1-2B              & 67.07          & 39.96          & 72.73          & \underline{86.76}          & 36.23          & 75.27          & 37.67          & 69.44          & 26.74          & 87.10           & 86.67           & 80.00           & 30.48          & \underline{85.71}          &\underline{83.33}           & 47.03                   \\
MedGemma-4B               & 71.22          & 45.36          & 72.73          & 62.62          & 40.94          & 63.44          & 43.74          & 77.78          & 28.00          & 87.10           & 80.00           & 60.00           & 26.03          & \textbf{92.86} &\underline{83.33}           & 48.04                   \\
Llava-Med-7B              & 54.35          & 33.91          & 63.64          & 65.10          & 26.09          & 51.61          & 33.51          & 52.78          & 22.53          & 48.39           & 60.00           & 30.00           & 27.62          & 42.86          & 66.67           & 38.84                   \\
Lingshu-7B                & 81.39          & 63.50          & \underline{77.27}         & 68.32          & 41.30          & 73.12          & 51.48          & 83.33          & 28.42          & \underline{96.77}           & 86.67           & 80.00           & 30.79          & 78.57          & 66.67           & 55.53                   \\
Lingshu-32B               & 60.51          & 61.77          & 72.73          & 60.40          & 40.94          & 73.12          & 55.13          & \underline{88.89}          & 31.37          & \underline{96.77}           & 73.33           & 80.00           & 33.65          & \underline{85.71}          & 66.67           & 54.39                   \\
HuatuoGPT-V-7B       & 82.06          & 66.74          & \underline{77.27}         & 82.43          & 45.65          & 78.49          & 54.58          & 83.33          & 28.84          & 90.32           & 86.67           & \underline{90.00}           & 31.11          & 64.29          & \textbf{100.00} & 59.37                   \\
HuatuoGPT-V-34B      & \underline{85.14}          & \textbf{69.55} & 72.73          & 84.41          & 48.19          & 78.49          & 58.68          & 80.56          & \textbf{40.84} & \underline{96.77}           & 86.67           & \underline{90.00}           & 39.05          & \underline{85.71}          & 66.67           & 63.56                   \\ 
\rowcolor{Purple!20} 
\multicolumn{17}{c}{\textbf{Proprietary MLLMs}} \\ 
Deepseek-V3.1             & 25.03          & 29.81          & 68.18          & 19.80          & 26.45          & 46.24          & 51.45          & 77.78          & 25.47          & 90.32           & 80.00           & 80.00           & 32.70          & \underline{85.71}          & 66.67           & 40.14                   \\
Grok-4                    & 72.29          & 48.38          & \textbf{81.82} & 65.84          & 36.96          & 69.89          & 57.86          & \underline{88.89}          & 30.74          & \underline{96.77}           & 86.67           & 80.00           & 34.29          & 78.57          & 66.67           & 56.65                   \\
GPT-4o                    & 77.91          & 65.66          & 72.73          & 72.77          & 48.91          & 77.42          & 60.14          & \textbf{91.67} & 38.11          & 93.55           & 73.33           & \underline{90.00}           & 37.46          & \underline{85.71}          & 66.67           & 61.64                   \\
GPT-5            & 80.59          & 65.87          & \textbf{81.82} & 74.88          & 48.91          & \underline{83.87}          & 61.73          & \textbf{91.67} & 38.11          & 93.55           & \textbf{100.00} & 80.00           & 41.27          & \textbf{92.86} & 66.67           & 63.37                   \\
GPT-5-mini        & 80.05          & 67.82          & 72.73          & 81.31          & \underline{51.45}          & 77.42          & \underline{62.73}          & \textbf{91.67} & \textbf{40.84} & \underline{96.77}           & 86.67           & 80.00           & \underline{44.76}          & 71.43          & 66.67           & \underline{65.00}                   \\
Claude-4.5-Sonnet & 80.46          & 65.87          & \textbf{81.82} & 70.67          & 46.01          & 79.57          & 57.46          & \textbf{91.67} & 31.79          & 93.55           & 80.00           & \textbf{100.00} & 36.51          & 78.57          & 66.67           & 59.78                   \\
Gemini-2.5-Pro            & \textbf{86.21} & \underline{69.11}          & \textbf{81.82} & 80.32          & \textbf{53.26} & 80.65          & \textbf{64.09} & \underline{88.89}          & \underline{40.63}          & 93.55           & 80.00           & \underline{90.00}           & \textbf{46.98} & \underline{85.71}          & 66.67           & \textbf{66.58}          \\ 
\hline\hline
\end{tabular}
}
\label{tab:closedendedqa}
\end{table*}

\subsection{Multiple Clinical Tasks}
\label{subsec:multitasks}
Existing brain imaging benchmarks \cite{labella2023asnr,moawad2024brain,baid2021rsna,hernandez2022isles,tbkk-q937-25,feczko2025baby} naive focus on limited scenarios, e.g., modality identification or organ classification, making evaluations insufficiently comprehensive to handle diverse clinical scenarios in real practice. To address this gap,~\ourbench~encompasses 15 specialized tasks within five primary phases:

\begin{itemize}
    \item \noindent\textbf{AIA establishes a framework for interpreting data, addressing \textit{``What are we looking at?''}}:
Anatomical Structure Identification (ASI) identifies normal organs and tissues in images.
Imaging Modality Identification (IMI) confirms the imaging modality source.
Anatomical Function Understanding (AFU) links structures to physiological functions.

    \item \textbf{LIL detects abnormal signals, addressing \textit{"Where is the abnormality and what does it look like?" }}:
Abnormal Signal Screening (AS) detects abnormal data.
Lesion Feature Description (LFD) details the lesion's morphology, size, and density.
Lesion Localization (LL) pinpoints the anatomical location of abnormalities.

    \item \textbf{DSCR integrates lesion data with medical knowledge for \textit{``What is this disease and why does it occur?''}}:
Disease Diagnosis Reasoning (DDR) combines data for diagnosis.
Pathophysiological Mechanism Correlation (PMC) links lesion traits to disease mechanisms.

    \item \textbf{PJRF assesses disease trajectory, addressing \textit{``What will happen?''}:}
Risk Stratification (RS) assigns risk levels based on diagnosis and condition.
Prognostic Factor Analysis (PFA) identifies factors affecting outcomes.
Clinical Sign Prediction (CSP) forecasts new symptoms in disease progression.
Drug Response Prediction (DRP) estimates drug treatment.

    \item \textbf{TCM creates a ``decision-execution-evaluation'' loop for treatment:}
Preoperative Assessment (PA) assesses risks before treatment decisions.
Treatment Plan Selection (TPS) sets the treatment strategy.
Postoperative Outcome Assessment (POA) evaluates efficacy and feedback.

\end{itemize}

\section{Experiments and Analysis}

\subsection{Experiment Setup}
We evaluate 24 MLLMs, comprising 10 open-source general-purpose models, 7 open-source medical-specific models, and 7 proprietary models accessed via APIs. 
For open-source general-purpose MLLMs, we assess Janus-Pro-7B~\cite{chen2025janus}, InternVL3-8B/9B/14B/38B~\cite{zhu2025internvl3}, Qwen2.5VL-7B/32B~\cite{qwen2.5-VL} and Qwen3VL-4B/8B/30B~\cite{qwen3-VL}, ranging from 7B to 38B. 
For open-source medical MLLMs, we assess MedVLM-R1-2B \cite{pan2025medvlm}, MedGemma-4B \cite{sellergren2025medgemma}, Llava-Med-7B \cite{li2023llava}, Lingshu-7B/32B~\cite{xu2025lingshu}, and HuatuoGPT-Vision-7B/34B~\cite{chen2024huatuogpt}. For proprietary MLLMs, we evaluate Deepseek-V3.1 \cite{deepseekai2024deepseekv3technicalreport,guo2025deepseek}, Grok-4\cite{grok4_2025}, GPT-4o, GPT-5 (08/07), GPT-5-mini (08/07)~\cite{openai2025gpt5}, Claude-4.5-Sonnet (09/29)~\cite{anthropic2025claude45}, and Gemini-2.5-Pro \cite{comanici2025gemini}. 
All experiments of open-source MLLMs are conducted with four NVIDIA A6000 GPUs.

\subsection{Evaluation Metrics} 
For closed-ended evaluations, model performance is assessed by computing accuracy as the proportion of exact matches between predicted outputs and ground-truth answers. For open-ended evaluations, we utilize a variety of metrics that provide different insights into model performance, including ROUGE1, ROUGEL \cite{lin2004rouge}, BLEU \cite{papineni2002bleu}, and BERTScore \cite{Zhang2020BERTScore}. 

\subsection{Experimental Results}
{We conduct extensive experiments on closed-ended VQA of~\ourbench~across five specialized clinical tasks with 15 secondary subtasks to summarize the capabilities and limitations of MLLMs.
From Table~\ref{tab:closedendedqa}, three key insights have been deduced as follows:}

\begin{itemize}
    \item {\textbf{Brain imaging analysis is challenging for MLLMs, with significant gaps between MLLMs and physicians.} Physicians achieve an average accuracy of 91.35\% across all tasks, whereas the highest-performing model, Gemini-2.5-Pro, attained only 66.58\%, reflecting a substantial performance gap of approximately 24.77\%. This disparity underscores the intrinsic complexity of brain imaging analysis, which necessitates both precise visual interpretation and specialized clinical expertise. It indicates that, while open-source models benefit from structured contextual inputs, they exhibit limitations in knowledge-intensive and reasoning-dependent domains, highlighting the critical need for domain-specific pretraining and reasoning capabilities.}
    \item {\textbf{Medical MLLMs exhibit heterogeneous performance.} The highest-performing HuatuoGPT-V-34B achieves a mean accuracy of 63.56\%, rendering it competitive with leading proprietary MLLMs, where it demonstrates superior performance in the clinical phases of IMI (69.55\%) and RS (40.84\%). In contrast, other medical MLLMs, e.g., MedGemma-4B (48.04\%) and Llava-Med-7B (38.84\%), display markedly lower aggregate scores, consistent with the observed general performance deficit. This suggests that while conducting domain-specific training, greater attention should be paid to balancing model generalization and task adaptability.}
    \item {\textbf{MLLMs expose the variation in task difficulty, exposing a gap between visual perception and medical comprehension.} MLLMs and physicians consistently achieve high scores in tasks like prognostic factor analysis, clinical sign prediction, drug response prediction, and postoperative outcome assessment, where perfect scores of 100.00\% are seen. Conversely, tasks like risk stratification and preoperative assessment appear much more difficult, with significantly lower scores across all MLLMs (e.g., the highest-performing MLLM scores 40.84\% in risk stratification). Our findings highlight the importance of integrating medical knowledge and clinical reasoning beyond visual perception to bridge the performance gap in complex diagnostic and decision-making tasks.}
\end{itemize}


\begin{table}[]
\centering
\caption{Performance of different MLLMs on open-ended VQA of~\ourbench. Higher values indicate better performance in generation quality, semantic similarity, and fluency.}
\resizebox{0.92\linewidth}{!}{
\begin{tabular}{l|cccc}
\hline\hline
Model   Name      & ROUGE1         & ROUGEL         & BLEU          & BERTScore     \\
\hline
\rowcolor{OrangeRed!20} 
\multicolumn{5}{c}{\textbf{Open-Source General-Purpose MLLMs}} \\ 
Janus-Pro-7B      & 8.00           & 5.75           & 0.62          & -20.96        \\
InternVL3-8B      & 22.43          & 13.84          & 1.47          & 5.68          \\
InternVL3-9B      & 13.81          & 9.16           & 0.75          & -7.78         \\
InternVL3-14B     & 20.67          & 13.57          & 1.29          & 5.53          \\
Qwen2.5-VL-7B     & 19.77          & 13.11          & 1.32          & 1.78          \\
Qwen2.5-VL-32B    & 15.60          & 9.77           & 0.89          & -3.07         \\
Qwen3-VL-4B       & 16.27          & 10.85          & 1.05          & -2.43         \\
Qwen3-VL-8B       & 20.19          & 12.68          & 1.23          & 2.12          \\
Qwen3-VL-30B      & 25.31          & \textbf{16.13} & 1.77          & 8.27          \\
\rowcolor{magenta!20} 
\multicolumn{5}{c}{\textbf{Open-Source Medical MLLMs}} \\ 
MedVLM-R1-2B      & 20.58          & 14.03          & 1.38          & 0.83          \\
MedGemma-4B       & 9.78           & 6.86           & 0.45          & -11.25        \\
Llava-Med-7B      & 10.59          & 6.83           & 0.47          & -47.21        \\
Lingshu-7B        & \underline{25.62} & 15.94          & \underline{1.88} & 8.47          \\
Lingshu-32B       & \textbf{26.45} & \underline{16.08} & \textbf{1.90} & 8.35          \\
HuatuoGPT-V-7B    & 20.43          & 13.01          & 1.33          & 0.52          \\
HuatuoGPT-V-34B   & 25.17          & 15.66          & 1.75          & 5.94          \\
\rowcolor{Purple!20} 
\multicolumn{5}{c}{\textbf{Proprietary MLLMs}} \\ 
Deepseek-V3.1     & 19.96          & 11.91          & 1.27          & 0.52          \\
Grok-4            & 21.53          & 11.45          & 1.19          & 4.07          \\
GPT-4o            & 23.81          & 14.28          & 1.55          & 3.88          \\
GPT-5             & 24.65          & 14.62          & 1.62          & \underline{9.08} \\
GPT-5-mini        & 23.90          & 14.03          & 1.51          & \textbf{9.13} \\
Claude-4.5-Sonnet & 22.64          & 13.19          & 1.50          & 5.21          \\
Gemini-2.5-Pro    & 20.71          & 12.20          & 1.37          & 2.06    \\
\hline\hline
\end{tabular}
}
\label{tab:openendedqa}
\end{table}

Additionally, based on the results in Table~\ref{tab:openendedqa}, we have the following observations for open-ended evaluations:

\begin{itemize}
    \item \textbf{Lingshu series dominate open-source and overall leadership.} Lingshu-32B decisively outperforms the much larger HuatuoGPT-V-34B, dominating lexical precision, fluency, and semantic alignment across all key metrics. It indicates that targeted multimodal architecture and data-efficient training now deliver superior generation quality over sheer parameter scale, proving efficiency trumps size in real-world MLLM performance.
    \item \textbf{Open-source MLLMs exhibit far greater performance variance than their proprietary counterparts.} While trailblazers like Lingshu claim the top spots across ROUGE1, ROUGEL, and BERTScore, many others—especially medical variants—languish at the bottom, which indicates that the open ecosystem's rapid, decentralized innovation fuels both groundbreaking advances and pronounced instability in model quality.
    \item \textbf{Proprietary MLLMs are more balanced than open-source MLLMs.} Open-source MLLMs surpass proprietary ones in ROUGE1 and BLEU, demonstrating the higher language consistency and fluency and revealing a paradigm shift in efficiency and accessibility.
\end{itemize}


\begin{figure}[t]
  \centering
   \includegraphics[width=0.88\linewidth]{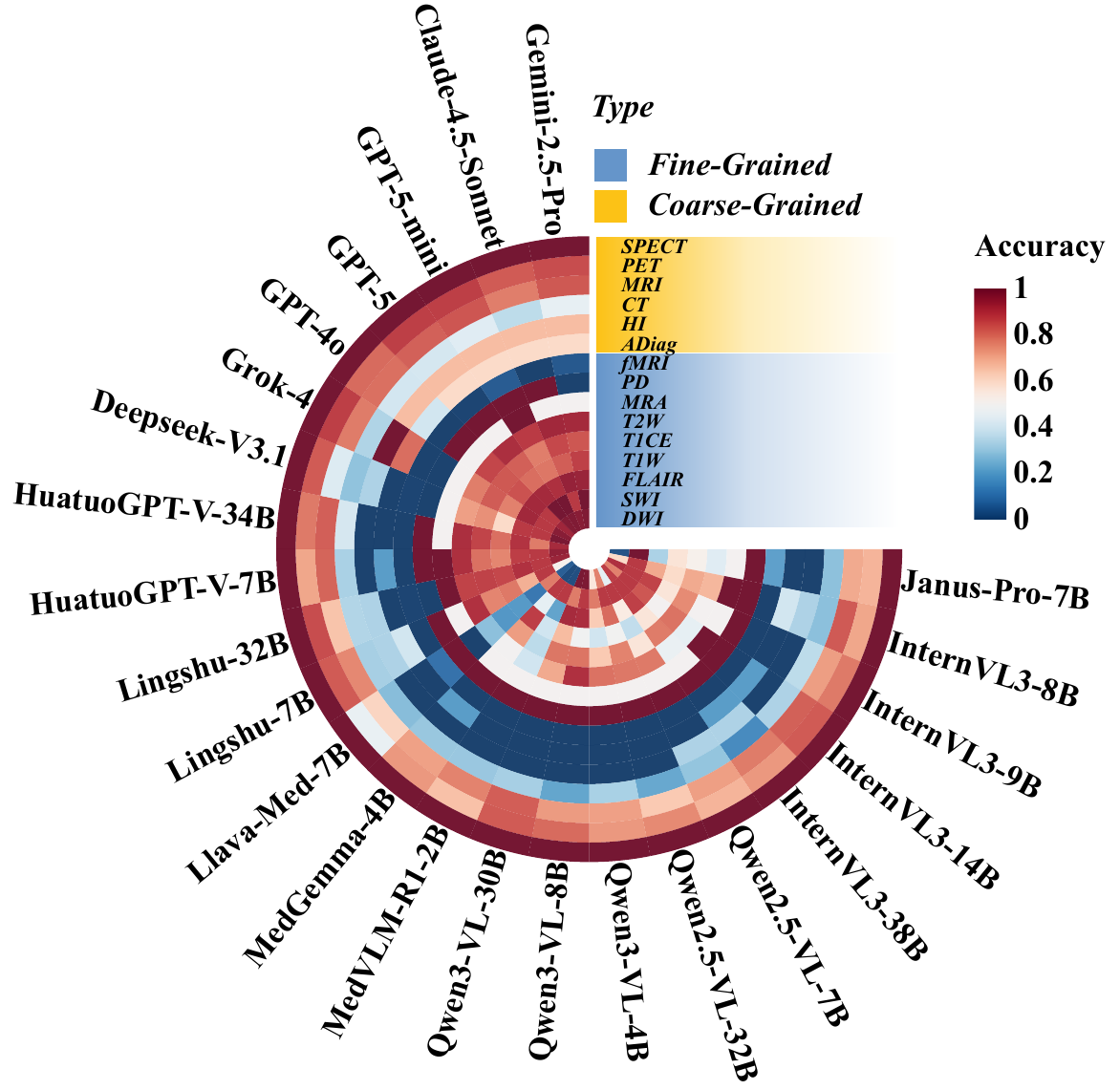}
   \caption{Diverse Modality Evaluation.}
   \label{fig: Diverse Modality Evaluation}
\end{figure}

\subsection{Diverse Modalities Analysis}
To better understand the modality-specific strengths and limitations of existing MLLMs, we conduct comparisons across 15 modalities in~\ourbench. From Fig.~\ref{fig: Diverse Modality Evaluation}, we can find that Gemini-2.5-Pro is the top generalist, but modality-specific strengths (e.g., Qwen3-VL-30B in FLAIR) highlight the value of targeted model selection. 
\begin{itemize}
    \item \textbf{Gemini-2.5-Pro leads in overall accuracy}: it shows strong performance across most modalities, with particularly high scores in PET (0.8537).
    \item \textbf{Large variation in modality-specific performance}: fMRI shows much lower performance across nearly all models ($\leq$ 0.5), indicating its challenging characteristics.
    \item \textbf{Smaller models can outperform larger ones in specific areas}: it suggests specialization or optimization in certain visual or medical imaging modalities.
\end{itemize}

\begin{figure}[t]
  \centering
   \includegraphics[width=0.98\linewidth]{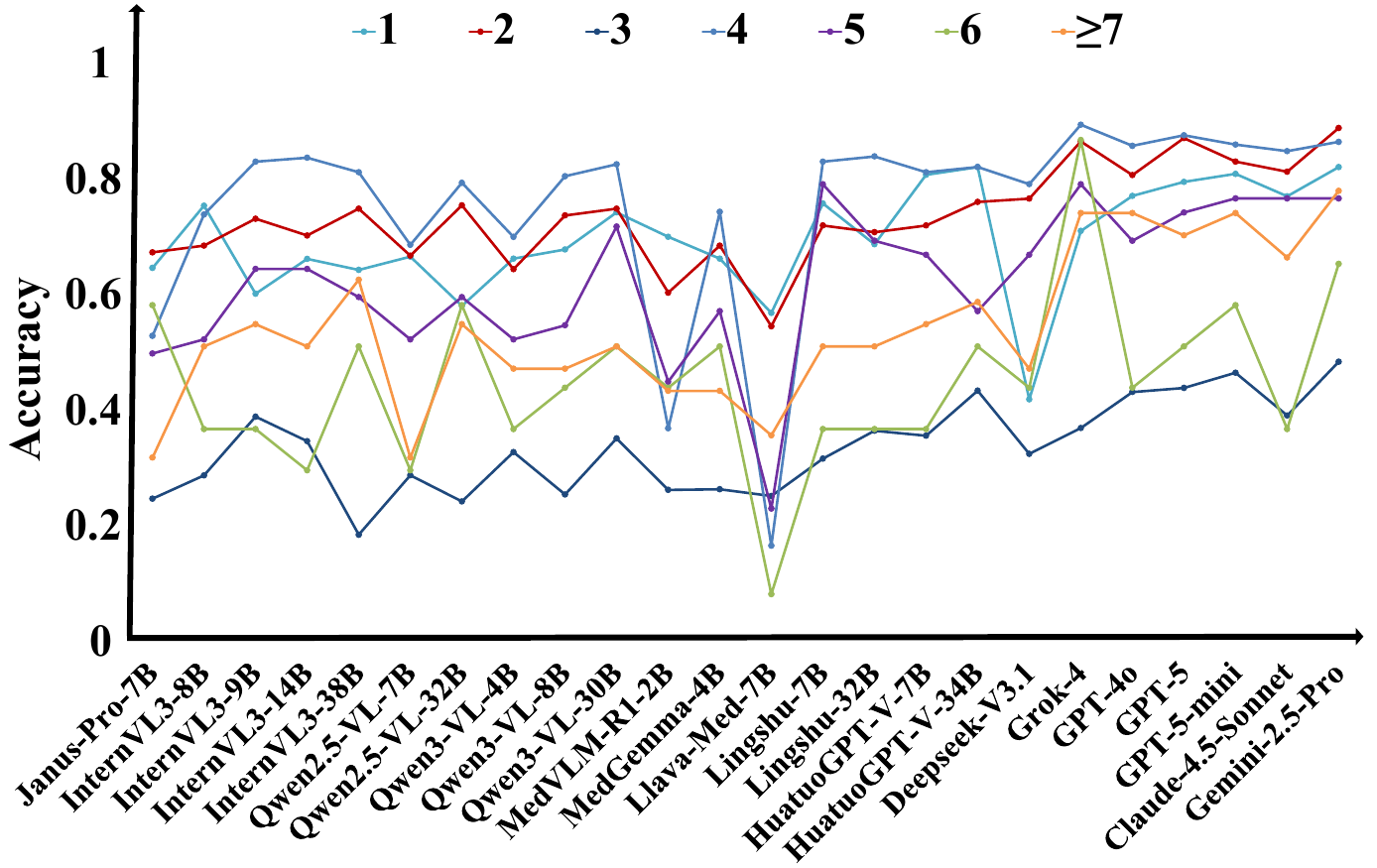}
   \caption{Performance of models on different numbers of images.}
   \label{fig:multiimg}
\end{figure}

\subsection{Multi-Image Analysis}
To better understand the scalability and robustness of existing MLLMs with respect to varying numbers of input images, we conduct performance comparisons on different numbers of images in~\ourbench, where models benefit significantly from 2–4 images, with Gemini-2.5-Pro excelling overall. From Fig. \ref{fig:multiimg}, we can find that:
\begin{itemize}
    \item \textbf{Gemini-2.5-Pro leads in overall accuracy and multi-image scaling}: it achieves the highest overall accuracy (66.58\%) and shows consistent gains with more images.
    \item \textbf{Performance generally improves with more images, peaking at four images}: it suggests strong multi-image reasoning capability, where models effectively integrate information from multiple visuals.
    \item \textbf{Diminishing or declining returns beyond four images}: it indicates potential information overload or reduced focus when too many images are provided.
\end{itemize}



\subsection{Discussion}
We aim to collaborate with the community to develop high-quality brain imaging benchmarks reflecting real clinical needs and align the model safety with human preferences. With these improvements, we plan real-world validation to assess practical efficacy. Crucially, while~\ourbench~is more comprehensive and clinically relevant than prior benchmarks, it cannot replace final clinical evaluation for safety. Instead, it serves as an experimental arena to accurately assess MLLM performance, reducing costs before expensive real-world deployments.

\section{Conclusion}
{We introduce~\ourbench, the comprehensive benchmark for evaluating MLLMs in brain imaging analysis, featuring both closed- and open-ended formats. Experimental results show that proprietary and medical MLLMs outperform their open-source counterparts across many tasks, while still lagging far behind human clinicians—especially in complex and nuanced scenarios. ~\ourbench~exposes critical gaps in clinical diagnosis and spatial reasoning, highlighting the need for advances in domain adaptation and prompt engineering. We expect~\ourbench~to drive progress toward clinically reliable MLLM for brain imaging.}




{
    \small
    \bibliographystyle{ieeenat_fullname}
    \bibliography{main}

\begin{thebibliography}{82}
\providecommand{\natexlab}[1]{#1}
\providecommand{\url}[1]{\texttt{#1}}
\expandafter\ifx\csname urlstyle\endcsname\relax
  \providecommand{\doi}[1]{doi: #1}\else
  \providecommand{\doi}{doi: \begingroup \urlstyle{rm}\Url}\fi

\bibitem[Abulnaga et~al.(2025)Abulnaga, Hoopes, Dey, Hoffmann, Fischl, Guttag, and Dalca]{abulnaga2025multimorph}
S~Mazdak Abulnaga, Andrew Hoopes, Neel Dey, Malte Hoffmann, Bruce Fischl, John Guttag, and Adrian Dalca.
\newblock Multimorph: On-demand atlas construction.
\newblock In \emph{Proceedings of the Computer Vision and Pattern Recognition Conference}, pages 30906--30917, 2025.

\bibitem[Alayrac et~al.(2022)Alayrac, Donahue, Luc, Miech, Barr, Hasson, Lenc, Mensch, Millican, Reynolds, et~al.]{alayrac2022flamingo}
Jean-Baptiste Alayrac, Jeff Donahue, Pauline Luc, Antoine Miech, Ivan Barr, Yana Hasson, Karel Lenc, Arthur Mensch, Katie Millican, Malcolm Reynolds, et~al.
\newblock Flamingo: A visual language model for few-shot learning.
\newblock In \emph{NeurIPS}, 2022.

\bibitem[Antel et~al.(2024)Antel, Ryvlin, et~al.]{eanm2024_epilepsy}
Jo{\~a}o Antel, Philippe Ryvlin, et~al.
\newblock Eanm practice guidelines for an appropriate use of pet and spect for epilepsy.
\newblock \emph{European Journal of Nuclear Medicine and Molecular Imaging}, 51:\penalty0 1315--1325, 2024.

\bibitem[{Anthropic}(2025)]{anthropic2025claude45}
{Anthropic}.
\newblock Claude sonnet 4.5, 2025.

\bibitem[Bai et~al.(2024)Bai, Dong, Cao, Zhao, Zhang, Wang, Ding, Chen, and Lv]{bai2024m3d}
Tianyu Bai, Yi Dong, Wentao Cao, Yunhao Zhao, Yanli Zhang, Jianmin Wang, Hui Ding, Feng Chen, and Kuanquan Lv.
\newblock M3d: Advancing 3d medical image analysis with multi-modal large language models.
\newblock \emph{arXiv preprint arXiv:2404.00511}, 2024.

\bibitem[Baid et~al.(2021)Baid, Ghodasara, Mohan, Bilello, Calabrese, Colak, Farahani, Kalpathy-Cramer, Kitamura, Pati, et~al.]{baid2021rsna}
Ujjwal Baid, Satyam Ghodasara, Suyash Mohan, Michel Bilello, Evan Calabrese, Errol Colak, Keyvan Farahani, Jayashree Kalpathy-Cramer, Felipe~C Kitamura, Sarthak Pati, et~al.
\newblock The rsna-asnr-miccai brats 2021 benchmark on brain tumor segmentation and radiogenomic classification.
\newblock \emph{arXiv preprint arXiv:2107.02314}, 2021.

\bibitem[Bercea et~al.(2025)Bercea, Li, Raffler, Riedel, Schmitzer, Kurz, Bitzer, Ro{\ss}m{\"u}ller, Canisius, Beyrle, et~al.]{bercea2025nova}
Cosmin~I Bercea, Jun Li, Philipp Raffler, Evamaria~O Riedel, Lena Schmitzer, Angela Kurz, Felix Bitzer, Paula Ro{\ss}m{\"u}ller, Julian Canisius, Mirjam~L Beyrle, et~al.
\newblock Nova: A benchmark for anomaly localization and clinical reasoning in brain mri.
\newblock \emph{arXiv preprint arXiv:2505.14064}, 2025.

\bibitem[Brown et~al.(2012)Brown, Sidhu, Greiner, Asgarian, Bastani, Silverstone, Greenshaw, and Dursun]{brown2012adhd}
Matthew~RG Brown, Gagan~S. Sidhu, Russell Greiner, Nasimeh Asgarian, Meysam Bastani, Peter~H. Silverstone, Andrew~J. Greenshaw, and Serdar~M. Dursun.
\newblock Adhd-200 global competition: diagnosing adhd using personal characteristic data can outperform resting state fmri measurements.
\newblock \emph{Frontiers in Systems Neuroscience}, 6:\penalty0 69, 2012.

\bibitem[Calhoun et~al.(2012)Calhoun, Sui, Kiehl, Turner, Allen, and Pearlson]{calhoun2012exploring}
Vince~D. Calhoun, Jing Sui, Kent Kiehl, Jessica Turner, Elena Allen, and Godfrey Pearlson.
\newblock Exploring the psychosis functional connectome: Aberrant intrinsic networks in schizophrenia and bipolar disorder.
\newblock \emph{Frontiers in Psychiatry}, 2:\penalty0 75, 2012.

\bibitem[Casey et~al.(2018)Casey, Cannonier, Conley, Cohen, Barch, Heitzeg, Soules, Teslovich, Dellarco, Garavan, et~al.]{casey2018adolescent}
Betty~Jo Casey, Tariq Cannonier, May~I Conley, Alexandra~O Cohen, Deanna~M Barch, Mary~M Heitzeg, Mary~E Soules, Theresa Teslovich, Danielle~V Dellarco, Hugh Garavan, et~al.
\newblock The adolescent brain cognitive development (abcd) study: imaging acquisition across 21 sites.
\newblock \emph{Developmental Cognitive Neuroscience}, 32:\penalty0 43--54, 2018.

\bibitem[Chang et~al.(2025)Chang, Lv, Guo, Lucia, Bollmann, Garner, McCombe, Henderson, Shaw, Steyn, et~al.]{chang2025fmri}
Jeryn Chang, JingLei Lv, Christine~C. Guo, Diana Lucia, Saskia Bollmann, Kelly Garner, Pamela~A. McCombe, Robert~D. Henderson, Thomas~B. Shaw, Frederik~J. Steyn, et~al.
\newblock An fmri dataset for appetite neural correlates in people living with motor neuron disease.
\newblock \emph{Scientific Data}, 12\penalty0 (1):\penalty0 466, 2025.

\bibitem[Chen et~al.(2024)Chen, Cai, Ji, Wang, Liu, Wang, Hou, and Wang]{chen2024huatuogpt}
Junying Chen, Zhenyang Cai, Ke Ji, Xidong Wang, Wanlong Liu, Rongsheng Wang, Jianye Hou, and Benyou Wang.
\newblock Huatuogpt-o1, towards medical complex reasoning with llms.
\newblock \emph{arXiv preprint arXiv:2412.18925}, 2024.

\bibitem[Chen et~al.(2025)Chen, Wu, Liu, Pan, Liu, Xie, Yu, and Ruan]{chen2025janus}
Xiaokang Chen, Zhiyu Wu, Xingchao Liu, Zizheng Pan, Wen Liu, Zhenda Xie, Xingkai Yu, and Chong Ruan.
\newblock Janus-pro: Unified multimodal understanding and generation with data and model scaling.
\newblock \emph{arXiv preprint arXiv:2501.17811}, 2025.

\bibitem[Chopra et~al.(2025)Chopra, Cocuzza, Lawhead, Ricard, Labache, Patrick, Kumar, Rubenstein, Moses, Chen, et~al.]{chopra2025transdiagnostic}
Sidhant Chopra, Carrisa~V Cocuzza, Connor Lawhead, Jocelyn~A Ricard, Lo{\"\i}c Labache, Lauren~M Patrick, Poornima Kumar, Arielle Rubenstein, Julia Moses, Lia Chen, et~al.
\newblock The transdiagnostic connectome project: an open dataset for studying brain-behavior relationships in psychiatry.
\newblock \emph{Scientific Data}, 12\penalty0 (1):\penalty0 923, 2025.

\bibitem[Comanici et~al.(2025)Comanici, Bieber, Schaekermann, Pasupat, Sachdeva, Dhillon, Blistein, Ram, Zhang, Rosen, et~al.]{comanici2025gemini}
Gheorghe Comanici, Eric Bieber, Mike Schaekermann, Ice Pasupat, Noveen Sachdeva, Inderjit Dhillon, Marcel Blistein, Ori Ram, Dan Zhang, Evan Rosen, et~al.
\newblock Gemini 2.5: Pushing the frontier with advanced reasoning, multimodality, long context, and next generation agentic capabilities.
\newblock \emph{arXiv preprint arXiv:2507.06261}, 2025.

\bibitem[consortium(2012)]{adhd2012adhd}
ADHD-200 consortium.
\newblock The adhd-200 consortium: a model to advance the translational potential of neuroimaging in clinical neuroscience.
\newblock \emph{Frontiers in systems neuroscience}, 6:\penalty0 62, 2012.

\bibitem[DeepSeek-AI(2024)]{deepseekai2024deepseekv3technicalreport}
DeepSeek-AI.
\newblock Deepseek-v3 technical report, 2024.

\bibitem[El~Zahraa El~Said et~al.(2023)El~Zahraa El~Said, Graves, and Lomas]{elzahraa2023multimodality}
Ahmed El~Zahraa El~Said, Martin~J Graves, and David~J Lomas.
\newblock Multimodality imaging of neurodegenerative disorders with a focus on multiparametric magnetic resonance and molecular imaging.
\newblock \emph{Insights into Imaging}, 14\penalty0 (1):\penalty0 27, 2023.

\bibitem[Feczko et~al.(2025)Feczko, Stoyell, Moore, Alexopoulos, Bagonis, Barrett, Bower, Cavender, Chamberlain, Conan, et~al.]{feczko2025baby}
Eric Feczko, Sally~M Stoyell, Lucille~A Moore, Dimitrios Alexopoulos, Maria Bagonis, Kenneth Barrett, Brad Bower, Addison Cavender, Taylor~A Chamberlain, Greg Conan, et~al.
\newblock Baby open brains: An open-source dataset of infant brain segmentations.
\newblock \emph{Scientific data}, 12\penalty0 (1):\penalty0 1423, 2025.

\bibitem[Guo et~al.(2025)Guo, Yang, Zhang, Song, Wang, Zhu, Xu, Zhang, Ma, Bi, et~al.]{guo2025deepseek}
Daya Guo, Dejian Yang, Haowei Zhang, Junxiao Song, Peiyi Wang, Qihao Zhu, Runxin Xu, Ruoyu Zhang, Shirong Ma, Xiao Bi, et~al.
\newblock Deepseek-r1 incentivizes reasoning in llms through reinforcement learning.
\newblock \emph{Nature}, 645\penalty0 (8081):\penalty0 633--638, 2025.

\bibitem[Gutbrod et~al.(2025)Gutbrod, Rauber, Nunes, and Palm]{gutbrod2025openmibood}
Max Gutbrod, David Rauber, Danilo~Weber Nunes, and Christoph Palm.
\newblock Openmibood: Open medical imaging benchmarks for out-of-distribution detection.
\newblock In \emph{Proceedings of the Computer Vision and Pattern Recognition Conference}, pages 25874--25886, 2025.

\bibitem[Hamada(2025)]{tbkk-q937-25}
Ahmed Hamada.
\newblock Br35h :: Brain tumor detection 2020, 2025.

\bibitem[Hernandez~Petzsche et~al.(2022)Hernandez~Petzsche, De~La~Rosa, Hanning, Wiest, Valenzuela, Reyes, Meyer, Liew, Kofler, Ezhov, et~al.]{hernandez2022isles}
Moritz~R Hernandez~Petzsche, Ezequiel De~La~Rosa, Uta Hanning, Roland Wiest, Waldo Valenzuela, Mauricio Reyes, Maria Meyer, Sook-Lei Liew, Florian Kofler, Ivan Ezhov, et~al.
\newblock Isles 2022: A multi-center magnetic resonance imaging stroke lesion segmentation dataset.
\newblock \emph{Scientific data}, 9\penalty0 (1):\penalty0 762, 2022.

\bibitem[Hu et~al.(2024)Hu, Li, Lu, Shao, He, Qiao, and Luo]{hu2024omnimedvqa}
Yutao Hu, Tianbin Li, Quanfeng Lu, Wenqi Shao, Junjun He, Yu Qiao, and Ping Luo.
\newblock Omnimedvqa: A new large-scale comprehensive evaluation benchmark for medical lvlm.
\newblock In \emph{Proceedings of the IEEE/CVF Conference on Computer Vision and Pattern Recognition}, pages 22170--22183, 2024.

\bibitem[Huang and Shu(2025)]{huang2025ai}
Weijie Huang and Ni Shu.
\newblock Ai-powered integration of multimodal imaging in precision medicine for neuropsychiatric disorders.
\newblock \emph{Cell Reports Medicine}, 6\penalty0 (5), 2025.

\bibitem[Irvin et~al.(2019)Irvin, Rajpurkar, Ko, et~al.]{irvin2019chexpert}
J. Irvin, P. Rajpurkar, M. Ko, et~al.
\newblock Chexpert: A large chest radiograph dataset with uncertainty labels and expert comparison.
\newblock \emph{Proceedings of the AAAI Conference on Artificial Intelligence}, 33\penalty0 (01):\penalty0 590--597, 2019.

\bibitem[Jack et~al.(2024)Jack, Veitch, Cash, Manning, Buckley, Raghavan, Schwarz, Tosun, Vemuri, Weiner, et~al.]{jack2024overview}
Clifford~R Jack, Dallas~P Veitch, David~M Cash, Emily Manning, Christopher Buckley, Nandini Raghavan, Adam~J Schwarz, Duygu Tosun, Prashanthi Vemuri, Michael~W Weiner, et~al.
\newblock Overview of adni mri.
\newblock \emph{Alzheimer's \& Dementia}, 2024.

\bibitem[Johnson et~al.(2019)Johnson, Pollard, Berkowitz, Greenbaum, Lungren, Deng, Mark, and Horng]{johnson2019mimic}
Alistair~EW Johnson, Tom~J Pollard, Seth~J Berkowitz, Nathaniel~R Greenbaum, Matthew~P Lungren, Chih-ying Deng, Roger~G Mark, and Steven Horng.
\newblock Mimic-cxr, a de-identified publicly available database of chest radiographs with free-text reports.
\newblock \emph{Scientific Data}, 6\penalty0 (1):\penalty0 317, 2019.

\bibitem[Kucs et~al.(2024)Kucs, Bohnsack, Szepesv{\'a}ri, et~al.]{kucs2024medsegbench}
Eszter Kucs, Nils Bohnsack, D{\'a}vid Szepesv{\'a}ri, et~al.
\newblock Medsegbench: A comprehensive benchmark for medical image segmentation in diverse data modalities.
\newblock \emph{Scientific Data}, 11\penalty0 (1):\penalty0 1234, 2024.

\bibitem[LaBella et~al.(2023)LaBella, Adewole, Alonso-Basanta, Altes, Anwar, Baid, Bergquist, Bhalerao, Chen, Chung, et~al.]{labella2023asnr}
Dominic LaBella, Maruf Adewole, Michelle Alonso-Basanta, Talissa Altes, Syed~Muhammad Anwar, Ujjwal Baid, Timothy Bergquist, Radhika Bhalerao, Sully Chen, Verena Chung, et~al.
\newblock The asnr-miccai brain tumor segmentation (brats) challenge 2023: Intracranial meningioma.
\newblock \emph{arXiv preprint arXiv:2305.07642}, 2023.

\bibitem[Lachinov et~al.(2023)Lachinov, Chakravarty, Grechenig, Schmidt-Erfurth, and Bogunovi{\'c}]{lachinov2023learning}
Dmitrii Lachinov, Arunava Chakravarty, Christoph Grechenig, Ursula Schmidt-Erfurth, and Hrvoje Bogunovi{\'c}.
\newblock Learning spatio-temporal model of disease progression with neuralodes from longitudinal volumetric data.
\newblock \emph{IEEE Transactions on Medical Imaging}, 43\penalty0 (3):\penalty0 1165--1179, 2023.

\bibitem[Lau et~al.(2018)Lau, Gayen, Ben~Abacha, and Demner-Fushman]{lau2018dataset}
Jason~J Lau, Soumya Gayen, Asma Ben~Abacha, and Dina Demner-Fushman.
\newblock A dataset of clinically generated visual questions and answers about radiology images.
\newblock \emph{Scientific data}, 5\penalty0 (1):\penalty0 1--10, 2018.

\bibitem[Li et~al.(2023{\natexlab{a}})Li, Wong, Zhang, Usuyama, Liu, Yang, Naumann, Poon, and Gao]{li2023llava}
Chunyuan Li, Cliff Wong, Sheng Zhang, Naoto Usuyama, Haotian Liu, Jianwei Yang, Tristan Naumann, Hoifung Poon, and Jianfeng Gao.
\newblock Llava-med: Training a large language-and-vision assistant for biomedicine in one day.
\newblock \emph{Advances in Neural Information Processing Systems}, 36:\penalty0 28541--28564, 2023{\natexlab{a}}.

\bibitem[Li et~al.(2022)Li, Li, Xiong, and Hoi]{li2022blip}
Junnan Li, Dongxu Li, Caiming Xiong, and Steven Hoi.
\newblock Blip: Bootstrapping language-image pre-training for unified vision-language understanding and generation.
\newblock In \emph{ICML}, 2022.

\bibitem[Li et~al.(2023{\natexlab{b}})Li, Li, Savarese, and Hoi]{li2023blip2}
Junnan Li, Dongxu Li, Silvio Savarese, and Steven Hoi.
\newblock Blip-2: Bootstrapping language-image pre-training with frozen image encoders and large language models.
\newblock In \emph{ICML}, 2023{\natexlab{b}}.

\bibitem[Lin(2004)]{lin2004rouge}
Chin-Yew Lin.
\newblock Rouge: A package for automatic evaluation of summaries.
\newblock In \emph{Text summarization branches out}, pages 74--81, 2004.

\bibitem[Liu et~al.(2021)Liu, Zhan, Xu, Ma, Yang, and Wu]{liu2021slake}
Bo Liu, Li-Ming Zhan, Li Xu, Lin Ma, Yan Yang, and Xiao-Ming Wu.
\newblock Slake: A semantically-labeled knowledge-enhanced dataset for medical visual question answering.
\newblock In \emph{2021 IEEE 18th International Symposium on Biomedical Imaging (ISBI)}, pages 1650--1654. IEEE, 2021.

\bibitem[Liu et~al.(2025{\natexlab{a}})Liu, Zheng, Chen, Peng, Yin, Shao, Hu, and Yuan]{liu2025comprehensive}
Shengyuan Liu, Boyun Zheng, Wenting Chen, Zhihao Peng, Zhenfei Yin, Jing Shao, Jiancong Hu, and Yixuan Yuan.
\newblock A comprehensive evaluation of multi-modal large language models for endoscopy analysis.
\newblock \emph{arXiv preprint arXiv:2505.23601}, 2025{\natexlab{a}}.

\bibitem[Liu et~al.(2025{\natexlab{b}})Liu, You, Chen, Zhang, Feng, Xu, Yu, and Cheng]{liu2025plasma}
Wei-Shi Liu, Jia You, Shi-Dong Chen, Yi Zhang, Jian-Feng Feng, Yu-Ming Xu, Jin-Tai Yu, and Wei Cheng.
\newblock Plasma proteomics identify biomarkers and undulating changes of brain aging.
\newblock \emph{Nature Aging}, 5\penalty0 (1):\penalty0 99--112, 2025{\natexlab{b}}.

\bibitem[Mahler et~al.(2020)Mahler, Tenney, and Provenzale]{mahler2020multimodality}
Simon Mahler, Nathan~D Tenney, and James~M Provenzale.
\newblock Multimodality imaging of dementia: Clinical importance and role of integrated anatomic and molecular imaging.
\newblock \emph{RadioGraphics}, 40\penalty0 (1):\penalty0 200--222, 2020.

\bibitem[Mei et~al.(2022)Mei, Liu, Robson, Marinelli, Huang, Doshi, Jacobi, Cao, Link, Yang, et~al.]{mei2022radimagenet}
Xueyan Mei, Zelong Liu, Philip~M Robson, Brett Marinelli, Mingqian Huang, Amish Doshi, Adam Jacobi, Chendi Cao, Katherine~E Link, Thomas Yang, et~al.
\newblock Radimagenet: an open radiologic deep learning research dataset for effective transfer learning.
\newblock \emph{Radiology: Artificial Intelligence}, 4\penalty0 (5):\penalty0 e210315, 2022.

\bibitem[Meling et~al.(2024)Meling, Egger, Aicher, Jare{\~n}o~Redondo, Mueller, Dornbierer, Temperli, Vasella, Caflisch, Pfeiffer, et~al.]{meling2024meditating}
Daniel Meling, Klemens Egger, Helena~D Aicher, Javier Jare{\~n}o~Redondo, Jovin Mueller, Jo{\"e}lle Dornbierer, Elijah Temperli, Emilia~A Vasella, Luzia Caflisch, David~J Pfeiffer, et~al.
\newblock Meditating on psychedelics. a randomized placebo-controlled study of dmt and harmine in a mindfulness retreat.
\newblock \emph{Journal of Psychopharmacology}, 38\penalty0 (10):\penalty0 897--910, 2024.

\bibitem[Menze et~al.(2015)Menze, Jakab, Bauer, Kalpathy-Cramer, Farahani, Kirby, Burren, Porz, Slotboom, Wiest, et~al.]{menze2015multimodal}
Bjoern~H Menze, Andras Jakab, Stefan Bauer, Jayashree Kalpathy-Cramer, Keyvan Farahani, Justin Kirby, Yuliya Burren, Nicole Porz, Johannes Slotboom, Roland Wiest, et~al.
\newblock The multimodal brain tumor image segmentation benchmark (brats).
\newblock \emph{IEEE Transactions on Medical Imaging}, 34\penalty0 (10):\penalty0 1993--2024, 2015.

\bibitem[Moawad et~al.(2024)Moawad, Janas, Baid, Ramakrishnan, Saluja, Ashraf, Maleki, Jekel, Yordanov, Fehringer, et~al.]{moawad2024brain}
Ahmed~W Moawad, Anastasia Janas, Ujjwal Baid, Divya Ramakrishnan, Rachit Saluja, Nader Ashraf, Nazanin Maleki, Leon Jekel, Nikolay Yordanov, Pascal Fehringer, et~al.
\newblock The brain tumor segmentation-metastases (brats-mets) challenge 2023: Brain metastasis segmentation on pre-treatment mri.
\newblock \emph{ArXiv}, pages arXiv--2306, 2024.

\bibitem[Mueller et~al.(2005)Mueller, Weiner, Thal, Petersen, Jack, Jagust, Trojanowski, Toga, and Beckett]{mueller2005alzheimer}
Susanne~G Mueller, Michael~W Weiner, Leon~J Thal, Ronald~C Petersen, Clifford Jack, William Jagust, John~Q Trojanowski, Arthur~W Toga, and Laurel Beckett.
\newblock The alzheimer's disease neuroimaging initiative.
\newblock \emph{Neuroimaging Clinics}, 15\penalty0 (4):\penalty0 869--877, 2005.

\bibitem[{New England Journal of Medicine}(2025)]{nejm2025imagechallenge}
{New England Journal of Medicine}.
\newblock Image challenge, 2025.

\bibitem[{OpenAI}(2025)]{openai2025gpt5}
{OpenAI}.
\newblock Gpt-5, 2025.

\bibitem[Oquab et~al.(2024)Oquab, Darcet, Moutakanni, Vo, Szafraniec, Khalidov, Fernandez, Haziza, Massa, El-Nouby, et~al.]{oquab2024dinov2}
Maxime Oquab, Timoth{\'e}e Darcet, Th{\'e}o Moutakanni, Huy Vo, Marc Szafraniec, Vasil Khalidov, Pierre Fernandez, Daniel Haziza, Francisco Massa, Alaaeldin El-Nouby, et~al.
\newblock Dinov2: Learning robust visual features without supervision.
\newblock \emph{Transactions on Machine Learning Research Journal}, pages 1--31, 2024.

\bibitem[Palmer(2007)]{palmer2007uk}
Lyle~J Palmer.
\newblock Uk biobank: bank on it.
\newblock \emph{The Lancet}, 369\penalty0 (9578):\penalty0 1980--1982, 2007.

\bibitem[Pan et~al.(2025)Pan, Liu, Wu, Liu, Zhu, Li, Chen, Ouyang, and Rueckert]{pan2025medvlm}
Jiazhen Pan, Che Liu, Junde Wu, Fenglin Liu, Jiayuan Zhu, Hongwei~Bran Li, Chen Chen, Cheng Ouyang, and Daniel Rueckert.
\newblock Medvlm-r1: Incentivizing medical reasoning capability of vision-language models (vlms) via reinforcement learning.
\newblock In \emph{International Conference on Medical Image Computing and Computer-Assisted Intervention}, pages 337--347. Springer, 2025.

\bibitem[Papineni et~al.(2002)Papineni, Roukos, Ward, and Zhu]{papineni2002bleu}
Kishore Papineni, Salim Roukos, Todd Ward, and Wei-Jing Zhu.
\newblock Bleu: a method for automatic evaluation of machine translation.
\newblock In \emph{Proceedings of the 40th annual meeting of the Association for Computational Linguistics}, pages 311--318, 2002.

\bibitem[Poldrack et~al.(2016)Poldrack, Congdon, Triplett, Gorgolewski, Karlsgodt, Mumford, Sabb, Freimer, London, Cannon, et~al.]{poldrack2016phenome}
Russell~A. Poldrack, Eliza Congdon, William Triplett, Krzysztof~J. Gorgolewski, Katherine~H. Karlsgodt, Jonathan~A. Mumford, Farrah~W. Sabb, Nelson~B. Freimer, Edythe~D. London, Tyrone~D. Cannon, et~al.
\newblock A phenome-wide examination of neural and cognitive function.
\newblock \emph{Scientific Data}, 3:\penalty0 160110, 2016.

\bibitem[Radford et~al.(2021)Radford, Kim, Hallacy, Ramesh, Goh, Agarwal, Sastry, Askell, Mishkin, Clark, et~al.]{radford2021clip}
Alec Radford, Jong~Wook Kim, Chris Hallacy, Aditya Ramesh, Gabriel Goh, Sandhini Agarwal, Girish Sastry, Amanda Askell, Pamela Mishkin, Jack Clark, et~al.
\newblock Learning transferable visual models from natural language supervision.
\newblock In \emph{ICML}, 2021.

\bibitem[{Radiopaedia contributors}(2024)]{radiopaedia}
{Radiopaedia contributors}.
\newblock Radiopaedia.
\newblock \url{https://radiopaedia.org}, 2024.
\newblock Accessed: 2024-06-10.

\bibitem[Ray et~al.(2024)Ray, Bade, Bhalerao, Agarwal, Kapil, Agarwal, Singh, Krishna, Rane, Mittal, Mitra, and Sheet]{ray2024nova}
Soumi Ray, Sairam Bade, Megh Bhalerao, Sumit Agarwal, Rudraksh Kapil, Anjali Agarwal, Sudhir~K Singh, Batchu~R Krishna, Swapnil Rane, Gaurav~S Mittal, Susmita Mitra, and Debdoot Sheet.
\newblock Nova: A benchmark for anomaly localization and clinical reasoning in brain mri.
\newblock \emph{arXiv preprint arXiv:2405.14064}, 2024.

\bibitem[Reimers and Gurevych(2019)]{reimers2019sentence}
Nils Reimers and Iryna Gurevych.
\newblock Sentence-bert: Sentence embeddings using siamese bert-networks.
\newblock In \emph{EMNLP-IJCNLP}, pages 3982--3992, 2019.

\bibitem[Seenivasan et~al.(2023)Seenivasan, Islam, Ren, and Mitheran]{seenivasan2022surgical}
Lalithkumar Seenivasan, Mobarakol Islam, Hongliang Ren, and Srinath Mitheran.
\newblock Surgical-vqla: Transformer with gated vision-language embedding for visual question localized-answering in robotic surgery.
\newblock In \emph{2023 IEEE International Conference on Robotics and Automation (ICRA)}, pages 6359--6365. IEEE, 2023.

\bibitem[Sellergren et~al.(2025)Sellergren, Kazemzadeh, Jaroensri, Kiraly, Traverse, Kohlberger, Xu, Jamil, Hughes, Lau, et~al.]{sellergren2025medgemma}
Andrew Sellergren, Sahar Kazemzadeh, Tiam Jaroensri, Atilla Kiraly, Madeleine Traverse, Timo Kohlberger, Shawn Xu, Fayaz Jamil, Cían Hughes, Charles Lau, et~al.
\newblock Medgemma technical report.
\newblock \emph{arXiv preprint arXiv:2507.05201}, 2025.

\bibitem[Shehzad et~al.(2025)Shehzad, Minutolo, Esposito, Fujita, and Aljuaid]{shehzad2025brain}
Faheem Shehzad, Aniello Minutolo, Massimo Esposito, Hamido Fujita, and Hanan Aljuaid.
\newblock Brain tumor mri interpretation: Towards a benchmark for medical visual question answering.
\newblock In \emph{International Conference on Industrial, Engineering and Other Applications of Applied Intelligent Systems}, pages 519--530. Springer, 2025.

\bibitem[Shin et~al.(2025)Shin, Lee, Jang, Son, Kim, and Hwang]{shin2025anatomical}
Yejee Shin, Yeeun Lee, Hanbyol Jang, Geonhui Son, Hyeongyu Kim, and Dosik Hwang.
\newblock Anatomical consistency and adaptive prior-informed transformation for multi-contrast mr image synthesis via diffusion model.
\newblock In \emph{Proceedings of the Computer Vision and Pattern Recognition Conference}, pages 30918--30927, 2025.

\bibitem[Stein et~al.(2019)Stein, Wu, Carr, Shih, Kalpathy-Cramer, Elliott, Prevedello, Kohli, Lungren, Culliton, Ball, and Halabi]{stein2019rsna}
Anouk Stein, Carol Wu, Chris Carr, George Shih, Jayashree Kalpathy-Cramer, Julia Elliott, Luciano Prevedello, Marc Kohli, Matt Lungren, Phil Culliton, Robyn Ball, and Safwan Halabi.
\newblock Rsna intracranial hemorrhage detection.
\newblock \url{https://kaggle.com/competitions/rsna-intracranial-hemorrhage-detection}, 2019.
\newblock Kaggle.

\bibitem[Tang et~al.(2024)Tang, Xiong, Tong, Yang, and Zhang]{tang2024multimodal}
Yan Tang, Xing Xiong, Gan Tong, Yuan Yang, and Hao Zhang.
\newblock Multimodal diagnosis model of alzheimer's disease based on improved transformer.
\newblock \emph{BioMedical Engineering OnLine}, 23\penalty0 (1):\penalty0 8, 2024.

\bibitem[Team(2025{\natexlab{a}})]{qwen2.5-VL}
Qwen Team.
\newblock Qwen2.5-vl, 2025{\natexlab{a}}.

\bibitem[Team(2025{\natexlab{b}})]{qwen3-VL}
Qwen Team.
\newblock Qwen3-vl: Sharper vision, deeper thought, broader action, 2025{\natexlab{b}}.

\bibitem[Thomalla et~al.(2018)Thomalla, Simonsen, Boutitie, Andersen, Berthezene, Cheng, Cheripelli, Cho, Fazekas, Fiehler, et~al.]{thomalla2018wake}
G{\"o}tz Thomalla, Claus~Z Simonsen, Florent Boutitie, Grethe Andersen, Yves Berthezene, Bo Cheng, Balaji Cheripelli, Tae-Hee Cho, Franz Fazekas, Jens Fiehler, et~al.
\newblock Mri-guided thrombolysis for stroke with unknown time of onset.
\newblock \emph{New England Journal of Medicine}, 379\penalty0 (7):\penalty0 611--622, 2018.

\bibitem[Van~Essen et~al.(2013)Van~Essen, Smith, Barch, Behrens, Yacoub, Ugurbil, Consortium, et~al.]{van2013wu}
David~C. Van~Essen, Stephen~M. Smith, Deanna~M. Barch, Timothy E.~J. Behrens, Essa Yacoub, Kamil Ugurbil, Wu-Minn~HCP Consortium, et~al.
\newblock The wu-minn human connectome project: an overview.
\newblock \emph{NeuroImage}, 80:\penalty0 62--79, 2013.

\bibitem[Wardlaw et~al.(2019)Wardlaw, Smith, and Dichgans]{wardlaw2019small}
Joanna~M Wardlaw, Colin Smith, and Martin Dichgans.
\newblock Small vessel disease: mechanisms and clinical implications.
\newblock \emph{The Lancet Neurology}, 18\penalty0 (7):\penalty0 684--696, 2019.

\bibitem[Wen et~al.(2010)Wen, Macdonald, Reardon, Cloughesy, Sorensen, Galanis, Degroot, Wick, Gilbert, Lassman, et~al.]{wen2010rano}
Patrick~Y Wen, David~R Macdonald, David~A Reardon, Timothy~F Cloughesy, Alma~G Sorensen, Evanthia Galanis, John Degroot, Wolfgang Wick, Mark~R Gilbert, Andrew~B Lassman, et~al.
\newblock Updated response assessment criteria for high-grade gliomas: Response assessment in neuro-oncology working group.
\newblock \emph{Journal of Clinical Oncology}, 28\penalty0 (11):\penalty0 1963--1972, 2010.

\bibitem[{xAI}(2025)]{grok4_2025}
{xAI}.
\newblock Grok-4.
\newblock \url{https://x.ai/news/grok-4}, 2025.

\bibitem[Xie et~al.(2024)Xie, Zhou, Gao, Wu, Li, Zhou, Liu, Xing, Zou, Xie, et~al.]{xie2024medtrinity}
Yunfei Xie, Ce Zhou, Lang Gao, Juncheng Wu, Xianhang Li, Hong-Yu Zhou, Sheng Liu, Lei Xing, James Zou, Cihang Xie, et~al.
\newblock Medtrinity-25m: A large-scale multimodal dataset with multigranular annotations for medicine.
\newblock \emph{arXiv preprint arXiv:2408.02900}, 2024.

\bibitem[Xu et~al.(2025)Xu, Chan, Li, Aljunied, Yuan, Wang, Xiao, Chen, Liu, Li, et~al.]{xu2025lingshu}
Weiwen Xu, Hou~Pong Chan, Long Li, Mahani Aljunied, Ruifeng Yuan, Jianyu Wang, Chenghao Xiao, Guizhen Chen, Chaoqun Liu, Zhaodonghui Li, et~al.
\newblock Lingshu: A generalist foundation model for unified multimodal medical understanding and reasoning.
\newblock \emph{arXiv preprint arXiv:2506.07044}, 2025.

\bibitem[Ye and Tang(2025)]{ye2025multimodal}
Jiarui Ye and Hao Tang.
\newblock Multimodal large language models for medicine: A comprehensive survey.
\newblock \emph{arXiv preprint arXiv:2504.21051}, 2025.

\bibitem[Yu et~al.(2025)Yu, Wang, Wu, Xie, and Zhou]{yu2025medframeqa}
Suhao Yu, Haojin Wang, Juncheng Wu, Cihang Xie, and Yuyin Zhou.
\newblock Medframeqa: A multi-image medical vqa benchmark for clinical reasoning.
\newblock \emph{arXiv preprint arXiv:2505.16964}, 2025.

\bibitem[Yue et~al.(2024)Yue, Ni, Zhang, Zheng, Liu, Zhang, Stevens, Jiang, Ren, Sun, et~al.]{yue2024mmmu}
Xiang Yue, Yuansheng Ni, Kai Zhang, Tianyu Zheng, Ruoqi Liu, Ge Zhang, Samuel Stevens, Dongfu Jiang, Weiming Ren, Yuxuan Sun, et~al.
\newblock Mmmu: A massive multi-discipline multimodal understanding and reasoning benchmark for expert agi.
\newblock In \emph{Proceedings of the IEEE/CVF Conference on Computer Vision and Pattern Recognition}, pages 9556--9567, 2024.

\bibitem[Yue et~al.(2025)Yue, Zheng, Ni, Wang, Zhang, Tong, Sun, Yu, Zhang, Sun, Su, Chen, and Neubig]{yue-etal-2025-mmmu}
Xiang Yue, Tianyu Zheng, Yuansheng Ni, Yubo Wang, Kai Zhang, Shengbang Tong, Yuxuan Sun, Botao Yu, Ge Zhang, Huan Sun, Yu Su, Wenhu Chen, and Graham Neubig.
\newblock {MMMU}-pro: A more robust multi-discipline multimodal understanding benchmark.
\newblock In \emph{Proceedings of the 63rd Annual Meeting of the Association for Computational Linguistics (Volume 1: Long Papers)}, Vienna, Austria, 2025. Association for Computational Linguistics.

\bibitem[Zbontar et~al.(2018)Zbontar, Knoll, Sriram, Murrell, Huang, Muckley, Defazio, Stern, Johnson, Bruno, et~al.]{zbontar2018fastmri}
Jure Zbontar, Florian Knoll, Anuroop Sriram, Tullie Murrell, Zhengnan Huang, Matthew~J Muckley, Aaron Defazio, Ruben Stern, Patricia Johnson, Mary Bruno, et~al.
\newblock fastmri: An open dataset and benchmarks for accelerated mri.
\newblock \emph{arXiv preprint arXiv:1811.08839}, 2018.

\bibitem[Zhang et~al.(2023{\natexlab{a}})Zhang, Tan, Han, Wang, and Meng]{zhang2023automatic}
Shujun Zhang, Liwei Tan, Qi Han, Hongyan Wang, and Jianli Meng.
\newblock Automatic report generation on a large-scale stroke mri dataset.
\newblock In \emph{2023 IEEE 6th International Conference on Electronic Information and Communication Technology (ICEICT)}, pages 123--128. IEEE, 2023{\natexlab{a}}.

\bibitem[Zhang* et~al.(2020)Zhang*, Kishore*, Wu*, Weinberger, and Artzi]{Zhang2020BERTScore}
Tianyi Zhang*, Varsha Kishore*, Felix Wu*, Kilian~Q. Weinberger, and Yoav Artzi.
\newblock Bertscore: Evaluating text generation with bert.
\newblock In \emph{International Conference on Learning Representations}, 2020.

\bibitem[Zhang et~al.(2023{\natexlab{b}})Zhang, Wu, Zhao, Lin, Zhang, Wang, and Xie]{zhang2023pmc}
Xiaoman Zhang, Chaoyi Wu, Ziheng Zhao, Weixiong Lin, Ya Zhang, Yanfeng Wang, and Weidi Xie.
\newblock Pmc-vqa: Visual instruction tuning for medical visual question answering.
\newblock \emph{arXiv preprint arXiv:2305.10415}, 2023{\natexlab{b}}.

\bibitem[Zhang et~al.(2025)Zhang, Lu, Ma, Zhang, Yue, and Sun]{zhang2025incomplete}
Zheyu Zhang, Yayuan Lu, Feipeng Ma, Yueyi Zhang, Huanjing Yue, and Xiaoyan Sun.
\newblock Incomplete multi-modal brain tumor segmentation via learnable sorting state space model.
\newblock In \emph{Proceedings of the Computer Vision and Pattern Recognition Conference}, pages 25982--25992, 2025.

\bibitem[Zhu et~al.(2025)Zhu, Wang, Chen, Liu, Ye, Gu, Tian, Duan, Su, Shao, et~al.]{zhu2025internvl3}
Jinguo Zhu, Weiyun Wang, Zhe Chen, Zhaoyang Liu, Shenglong Ye, Lixin Gu, Hao Tian, Yuchen Duan, Weijie Su, Jie Shao, et~al.
\newblock Internvl3: Exploring advanced training and test-time recipes for open-source multimodal models.
\newblock \emph{arXiv preprint arXiv:2504.10479}, 2025.

\bibitem[Zuo et~al.(2025)Zuo, Qu, Li, Chen, Zhu, Hua, Zhang, Ding, and Zhou]{zuo2025medxpertqa}
Yuxin Zuo, Shang Qu, Yifei Li, Zhang-Ren Chen, Xuekai Zhu, Ermo Hua, Kaiyan Zhang, Ning Ding, and Bowen Zhou.
\newblock Medxpert{QA}: Benchmarking expert-level medical reasoning and understanding.
\newblock In \emph{ICML}, 2025.

\end{thebibliography}
}


\appendix





\clearpage
\setcounter{page}{1}
\maketitlesupplementary





\section{Ethics Statement}

\noindent\textbf{Ethical use of brain imaging data.}~\ourbench~is developed with a strong commitment to ethical practices in handling brain imaging data. All data included in the benchmark are sourced from open-access repositories and published articles. The dataset has been rigorously anonymized and contains no patient-identifiable information, ensuring full compliance with applicable privacy and research ethics guidelines. This collection and benchmarking process does not constitute human subjects research.

\noindent\textbf{Potential societal impacts.} Both positive and negative effects on brain imaging study and clinical practice are possible with AI models.  Advanced MLLMs have the potential to improve scientific research discovery and speed up diagnostic procedures, but they also run the risk of reinforcing biases in training data, which could result in uneven performance across various demographic groups or neurological conditions.~\ourbench~uses structured metadata to support the analysis of model biases and fairness in order to help reduce these risks.  In order to proactively address these and other new ethical issues, we are dedicated to continuing to engage with the research community. 

\noindent\textbf{Data licensing and usage.}~\ourbench~is distributed under the Creative Commons Attribution-ShareAlike 4.0 International license (CC BY-SA 4.0). This licensing framework is chosen to promote transparency, collaboration, and the responsible open-sourcing of resources within the research community. It enables both academic and commercial applications of the benchmark while ensuring that subsequent adaptations and distributions adhere to the same open and ethical principles.

\begin{figure*}[t]
  \centering
   \includegraphics[width=0.66\linewidth]{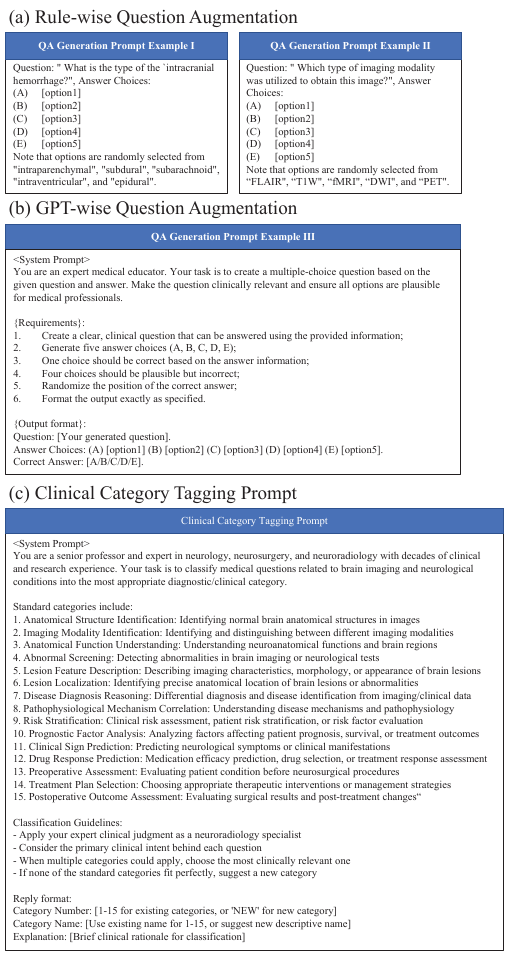}
   \caption{The prompt engineering is the process of designing and refining prompts (instructions) to guide generative AI models toward producing specific, high-quality outputs.}
   \label{fig: Prompts}
\end{figure*}

\section{Benchmark Construction}
To ensure a comprehensive and multifaceted evaluation of model performance, we develop a set of distinct question templates, as illustrated in Fig. \ref{fig: Prompts}. These templates are meticulously designed to systematically outline the specific prompts associated with each diagnostic task.

Specifically, our approach incorporates two complementary augmentation strategies: rule-wise question augmentation and GPT-wise question augmentation. The rule-based strategy employs structured templates with randomized answer choices, such as identifying hemorrhage types or imaging modalities, to ensure consistency and control over question formulation. In parallel, the GPT-based strategy leverages advanced language models guided by detailed system prompts to generate clinically relevant multiple-choice questions. These prompts require the generation of plausible distractors, randomization of the correct answer’s position, and strict adherence to a standardized output format, thereby enhancing clinical authenticity and variety. Furthermore, to support granular performance analysis across diagnostic subtasks, we introduced a clinical category tagging prompt mechanism. This allows each generated question to be classified into one of 15 predefined clinical categories. The mapping of questions to clinical domains is conducted under the supervision of a board-certified radiologist with over 13 years of experience, with GPT-5 performing the classification task. This process ensures that the assignments meet the highest standards of clinical relevance.

This structured framework facilitates the generation of a diverse and targeted question bank, encompassing a wide spectrum of clinical scenarios and varying levels of complexity. Consequently, it enables the robust and rigorous testing of MLLMs across the entire spectrum of brain imaging analysis capabilities, from foundational anatomical recognition to advanced clinical reasoning.

\begin{table}[]
\caption{Details of The Abbreviation}
\label{tab:abbr}
\begin{tabular}{ll}
\hline\hline
\multicolumn{2}{c}{Modality}                       \\
\hline
CT    & Computed Tomography                        \\
MRI   & magnetic resonance imaging                 \\
DWI   & diffusion-weighted imaging                 \\
T2W   & T2-weighted imaging                        \\
T1W   & T1-weighted imaging                        \\
FLAIR & fluid-attenuated inversion recovery        \\
MRA   & magnetic resonance angiography             \\
T1CE  & T1-weighted contrast-enhanced              \\
fMRI  & functional magnetic resonance imaging      \\
SWI   & susceptibility-weighted imaging            \\
PET   & positron emission tomography               \\
PD    & proton density weighted imaging            \\
HI    & histopathology imaging                     \\
ADiag & anatomical diagram                         \\
SPECT & single-photon emission computed tomography \\
\hline\hline
\multicolumn{2}{c}{Task}                           \\
\hline
AFU   & Anatomical Function Understanding          \\
AS    & Abnormal Screening                         \\
ASI   & Anatomical Structure Identification        \\
CSP   & Clinical Sign Prediction                   \\
DDR   & Disease Diagnosis Reasoning                \\
DRP   & Drug Response Prediction                   \\
IMI   & Imaging Modality Identification            \\
LFD   & Lesion Feature Description                 \\
LL    & Lesion Localization                        \\
PFA   & Prognostic Factor Analysis                 \\
PMC   & Pathophysiological Mechanism Correlation   \\
POA   & Postoperative Outcome Assessment           \\
PA    & Preoperative Assessment                    \\
RS    & Risk Stratification                        \\
TPS   & Treatment Plan Selection                   \\
AIA   & Anatomical and Imaging Assessment          \\
DSCR  & Diagnostic Synthesis and Causal Reasoning  \\
LIL   & Lesion Identification and Localization     \\
PJRF  & Prognostic Judgment and Risk Forecasting   \\
TCM   & Therapeutic Cycle Management               \\
\hline\hline
\end{tabular}
\end{table}

\section{Details of Abbreviation}
Brain imaging analysis relies on a diverse set of imaging modalities to visualize internal anatomy and function, which are interpreted through a hierarchy of clinical tasks to support diagnosis and treatment. The details of the abbreviations are given in Table \ref{tab:abbr}.

A wide array of modalities provides complementary information for clinical assessment. Cross-sectional imaging techniques like Computed Tomography (CT) and Magnetic Resonance Imaging (MRI) form the cornerstone. MRI itself encompasses numerous specialized sequences, each highlighting different tissue properties. These include T2-weighted (T2W) and T1-weighted (T1W) imaging for anatomical detail, Fluid-attenuated Inversion Recovery (FLAIR) for suppressing cerebrospinal fluid, and Diffusion-Weighted Imaging (DWI) for detecting cellular density. Further sequences like Magnetic Resonance Angiography (MRA) visualize vasculature, while T1-weighted Contrast-Enhanced (T1CE) imaging assesses vascular permeability and inflammation. Functional MRI (fMRI) maps brain activity, and Susceptibility-Weighted Imaging (SWI) is sensitive to blood products. In nuclear medicine, Positron Emission Tomography (PET) and Single-Photon Emission Computed Tomography (SPECT) provide metabolic and functional data. Proton Density Weighted (PD) imaging offers another contrast mechanism in MRI, and Histopathology Imaging (HI) remains the gold standard for definitive diagnosis. Anatomical Diagrams (ADiag) are often used for reference and education.

The tasks performed using these modalities can be categorized from foundational to advanced. The foundation begins with Imaging Modality Identification (IMI) and Anatomical Structure Identification (ASI), which are prerequisites for higher-level reasoning. Anatomical Function Understanding (AFU) builds upon this structural knowledge. The core of radiological analysis involves Abnormal Screening (AS), Lesion Localization (LL), and Lesion Feature Description (LFD). These can be grouped under the broader umbrella of Lesion Identification and Localization (LIL). dvanced tasks integrate these findings for clinical decision-making. This includes Clinical Sign Prediction (CSP), Disease Diagnosis Reasoning (DDR), and understanding the Pathophysiological Mechanism Correlation (PMC). Together, ASI, AFU, and IMI form the basis of a comprehensive Anatomical and Imaging Assessment (AIA), while DDR and PMC are key components of Diagnostic Synthesis and Causal Reasoning (DSCR). Management-focused tasks include Preoperative Assessment (PA), Treatment Plan Selection (TPS), and Prognostic Factor Analysis (PFA), which contributes to Prognostic Judgment and Risk Forecasting (PJRF). Risk Stratification (RS) is another critical prognostic task. Following intervention, Postoperative Outcome Assessment (POA) and Drug Response Prediction (DRP) are essential for monitoring, both falling under the scope of Therapeutic Cycle Management (TCM).

\section{Diverse Disease Coverage}

We have collaborated closely with board-certified radiologists to systematically categorize all diseases appearing in the dataset into two distinct groups: independent diseases and descriptive (non-independent) diseases.

On the one hand, independent diseases refer to well-defined, standalone clinical entities with specific histopathological, genetic, or etiological characteristics that allow them to be diagnosed as distinct nosological units. Examples from our dataset include Meningioma (the frequency count is 60), Glioblastoma (36), Pituitary Adenoma (24), Metastasis (19), Astrocytoma (12), Schwannoma (11), and rare but highly specific entities such as Dysplastic Cerebellar Gangliocytoma (1). These conditions typically present characteristic imaging features and are the final clinical diagnoses recorded in radiology reports.

On the other hand, descriptive diseases are descriptive pathological terms or imaging findings that do not constitute a final, standalone diagnosis but instead describe structural, developmental, or secondary abnormalities that may occur across a wide spectrum of underlying etiologies. Representative examples include Vermis and Midbrain Malformation, Agenesis of the Corpus Callosum, Cortical Dysplasia, Arachnoid Cyst, Chiari I Malformation, Leigh Syndrome, Periventricular Leukomalacia, Holoprosencephaly, Lissencephaly, and vascular anomalies such as Cavernous Malformation or Developmental Venous Anomaly. These findings are important radiological signs, but they usually require integration with clinical context and additional workup to reach a definitive diagnosis. A key motivation behind this fine-grained categorization and the construction of~\ourbench~stems from a critical limitation observed in existing public brain imaging benchmarks created for Disease Diagnosis and Reasoning (DDR) tasks. Most prior datasets and challenges predominantly focus on a handful of broad, nonspecific categories such as “Tumor”, “Stroke”, “Edema”, “Hemorrhage”, or “Normal/Mild Atrophy.” While these labels are clinically frequent, they fail to reflect the true complexity and diversity that radiologists and neurologists encounter in daily practice, where hundreds of rare and highly specific diagnoses must be considered in the differential.

As evidenced in Table \ref{tab:diverse disease} and the accompanying frequency distribution,~\ourbench~contains 218 unique, radiologist-verified diagnosis labels—an order of magnitude greater than the typical 8–20 classes found in existing benchmarks. Every single label in the dataset has been individually reviewed and validated by at least one senior neuroradiologist to ensure diagnostic accuracy and clinical meaningfulness. This rigorous annotation process guarantees that~\ourbench~not only dramatically exceeds prior benchmarks in breadth and depth of disease coverage but also provides a clinically authentic testing ground for evaluating the true diagnostic and reasoning capabilities of modern medical vision-language models. By forcing models to distinguish between subtle yet critical entities, such as differentiating a low-grade Dysplastic Cerebellar Gangliocytoma from a Medulloblastoma, or recognizing the characteristic imaging pattern of Leigh Syndrome versus hypoxic-ischemic injury, we establish a significantly more challenging and clinically representative benchmark for the DDR task.


\begin{figure*}[t]
  \centering
   \includegraphics[width=0.88\linewidth]{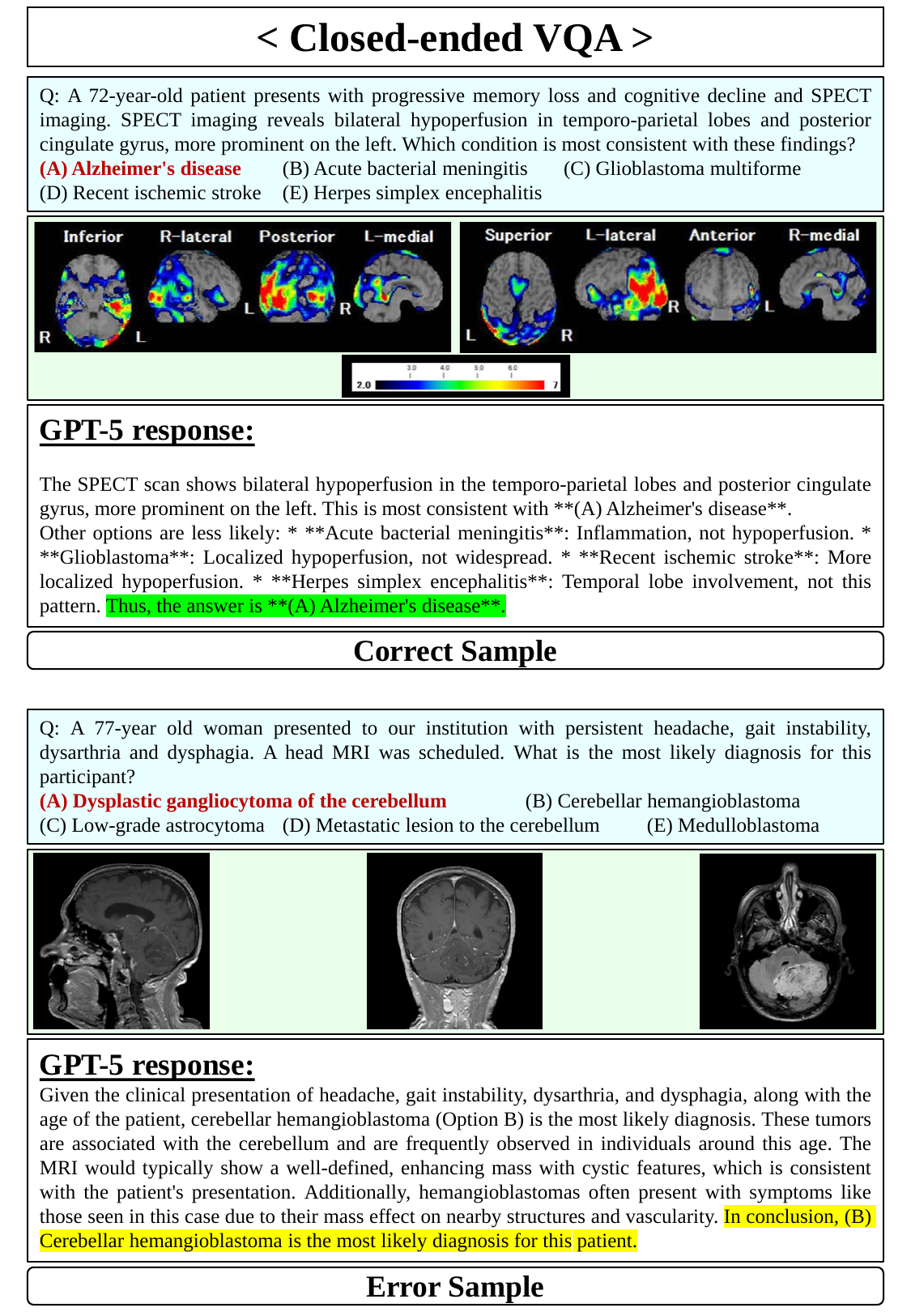}
   \caption{Correct/Error samples in GPT-5 closed-ended VQA.}
   \label{fig: GPT-5 closed-ended VQA}
\end{figure*}

\begin{figure*}[t]
  \centering
   \includegraphics[width=0.88\linewidth]{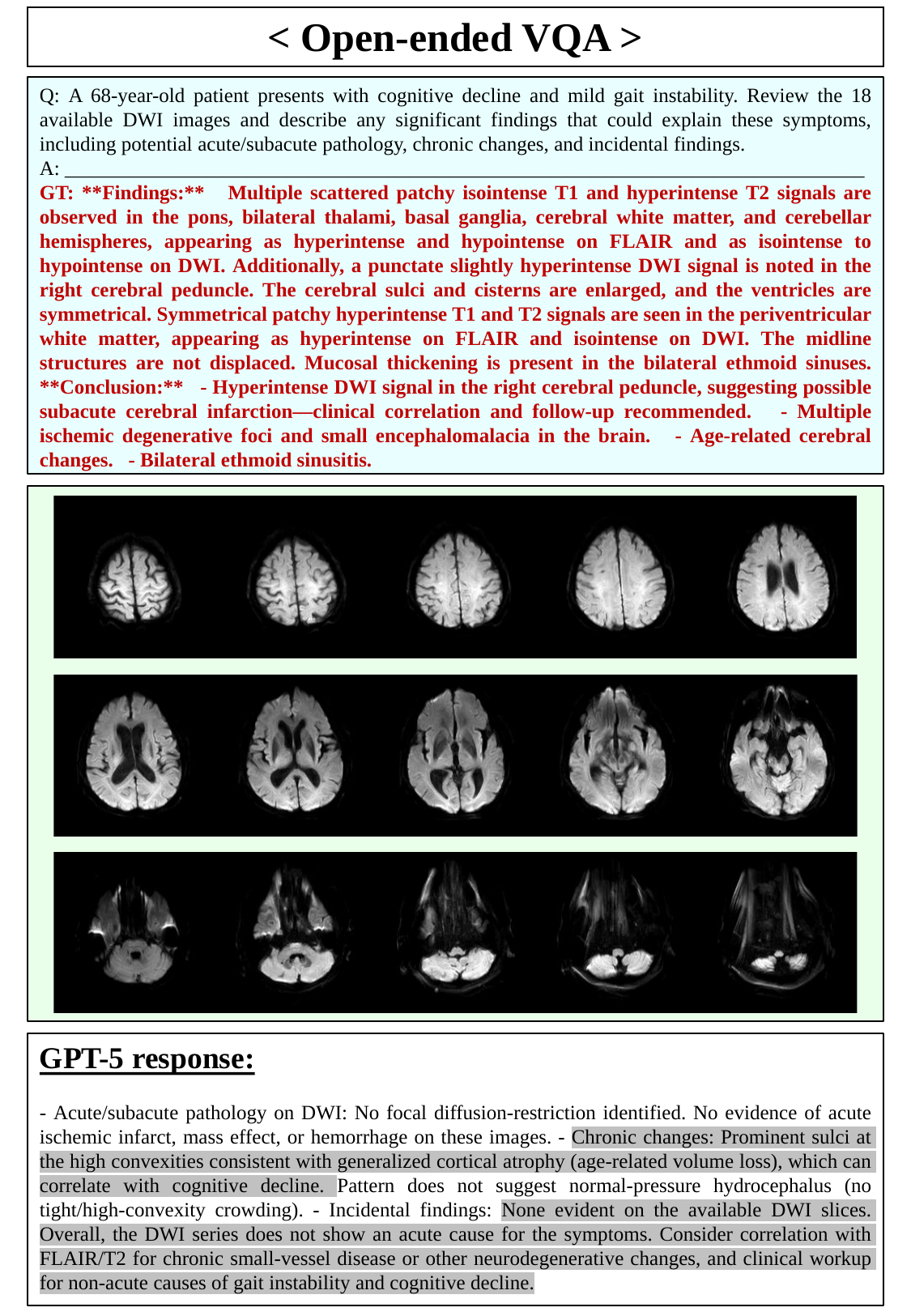}
   \caption{Correct/Error samples in GPT-5 open-ended VQA.}
   \label{fig: GPT-5 open-ended VQA}
\end{figure*}

\begin{figure*}[t]
  \centering
   \includegraphics[width=0.88\linewidth]{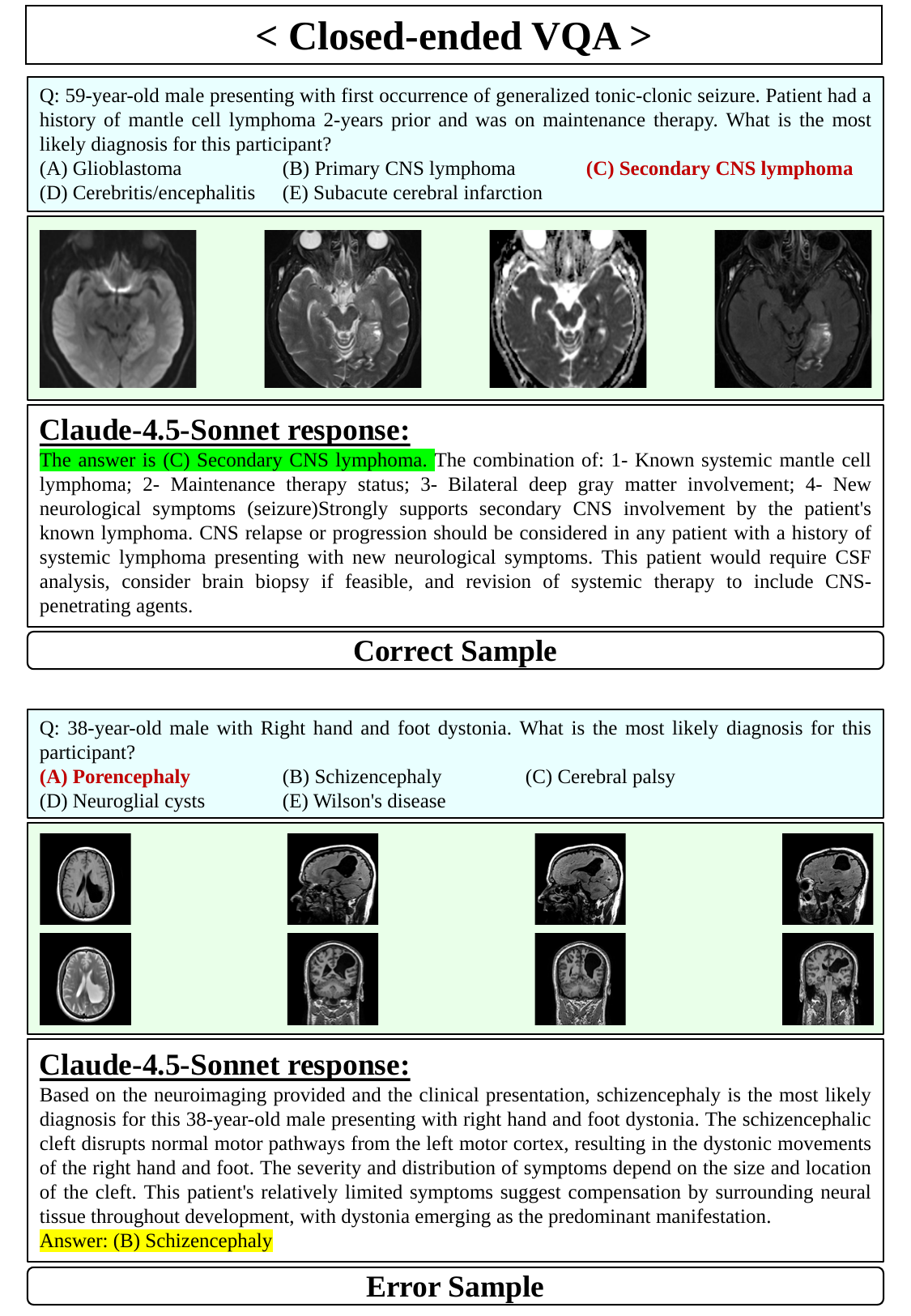}
   \caption{Correct/Error samples in Claude-4.5-Sonnet closed-ended VQA.}
   \label{fig: Claude-4.5-Sonnet closed-ended VQA}
\end{figure*}

\begin{figure*}[t]
  \centering
   \includegraphics[width=0.88\linewidth]{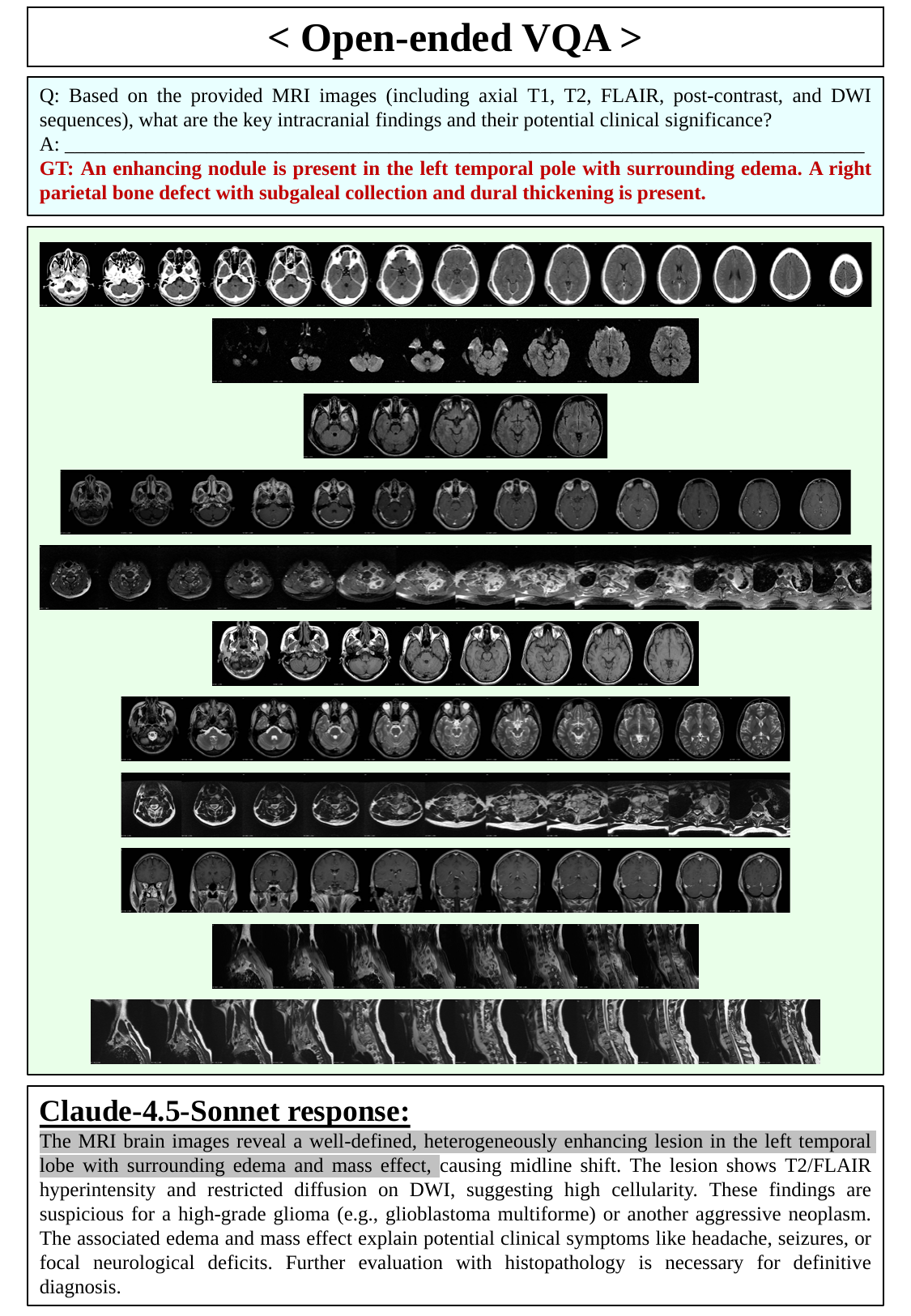}
   \caption{Correct/Error samples in Claude-4.5-Sonnet open-ended VQA.}
   \label{fig: Claude-4.5-Sonnet open-ended VQA}
\end{figure*}

\begin{figure*}[t]
  \centering
   \includegraphics[width=0.88\linewidth]{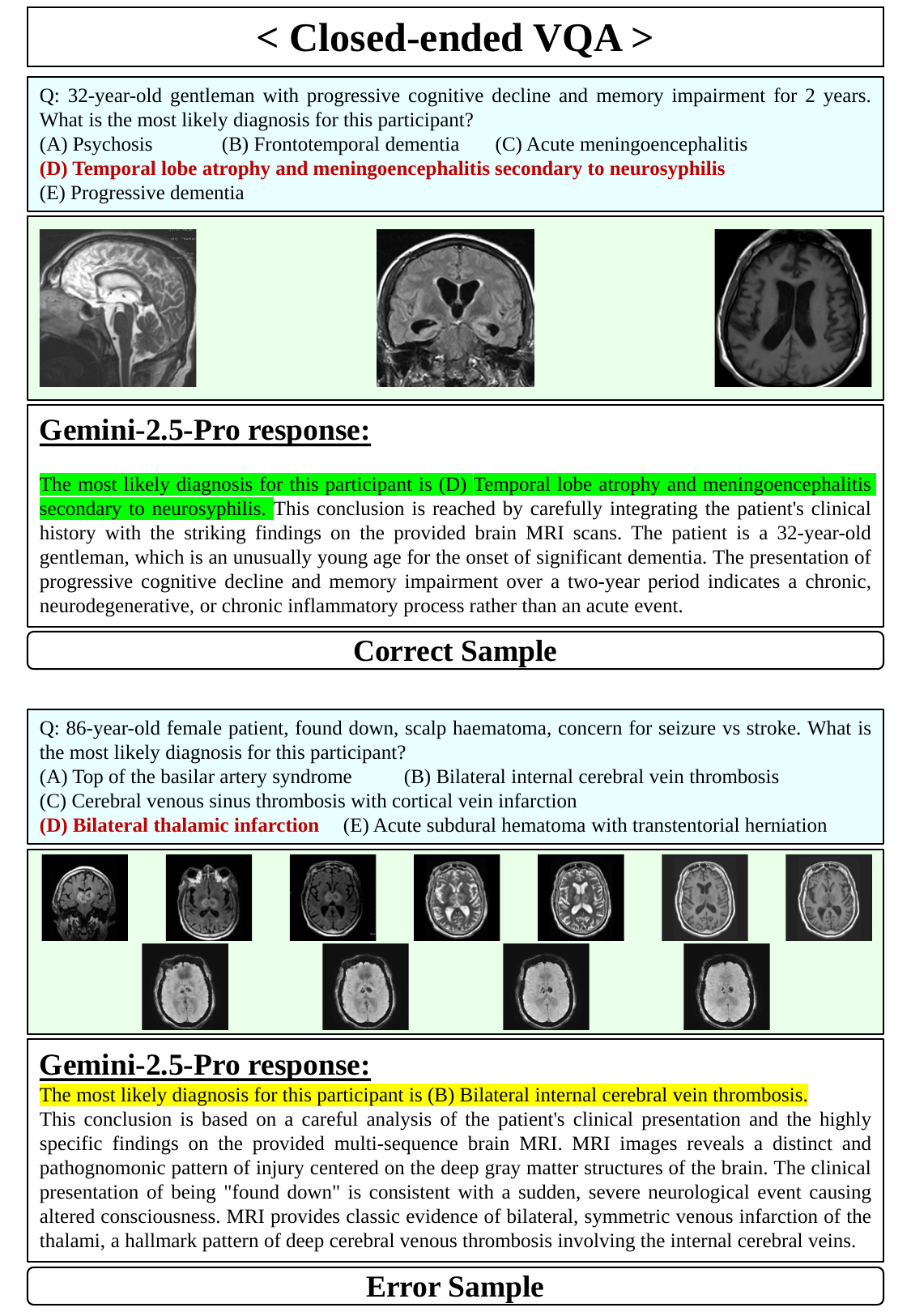}
   \caption{Correct/Error samples in Gemini-2.5-Pro closed-ended VQA.}
   \label{fig: Gemini-2.5-Pro closed-ended VQA}
\end{figure*}

\begin{figure*}[t]
  \centering
   \includegraphics[width=0.88\linewidth]{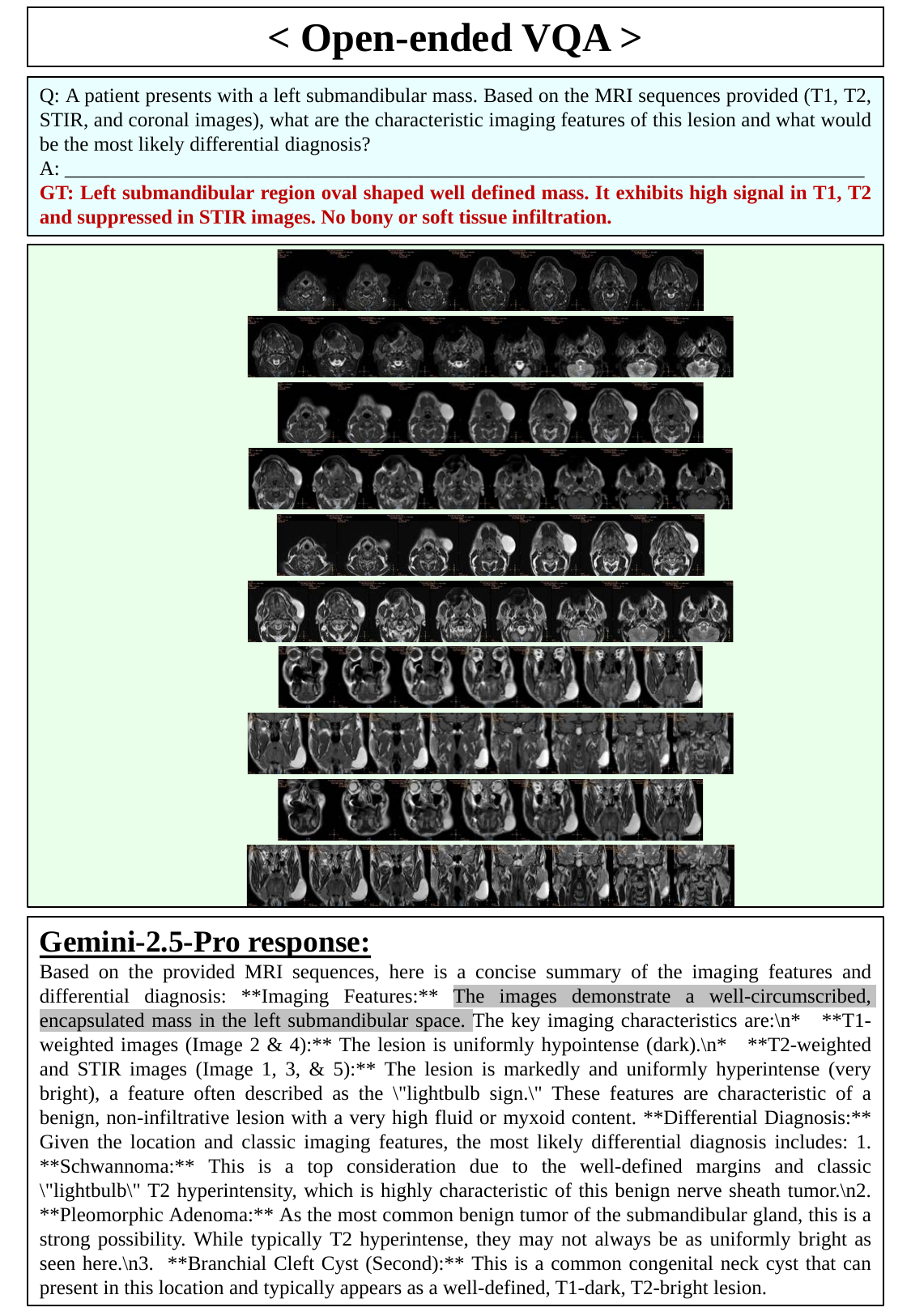}
   \caption{Correct/Error samples in Gemini-2.5-Pro open-ended VQA.}
   \label{fig: Gemini-2.5-Pro open-ended VQA}
\end{figure*}

\begin{figure*}[t]
  \centering
   \includegraphics[width=0.88\linewidth]{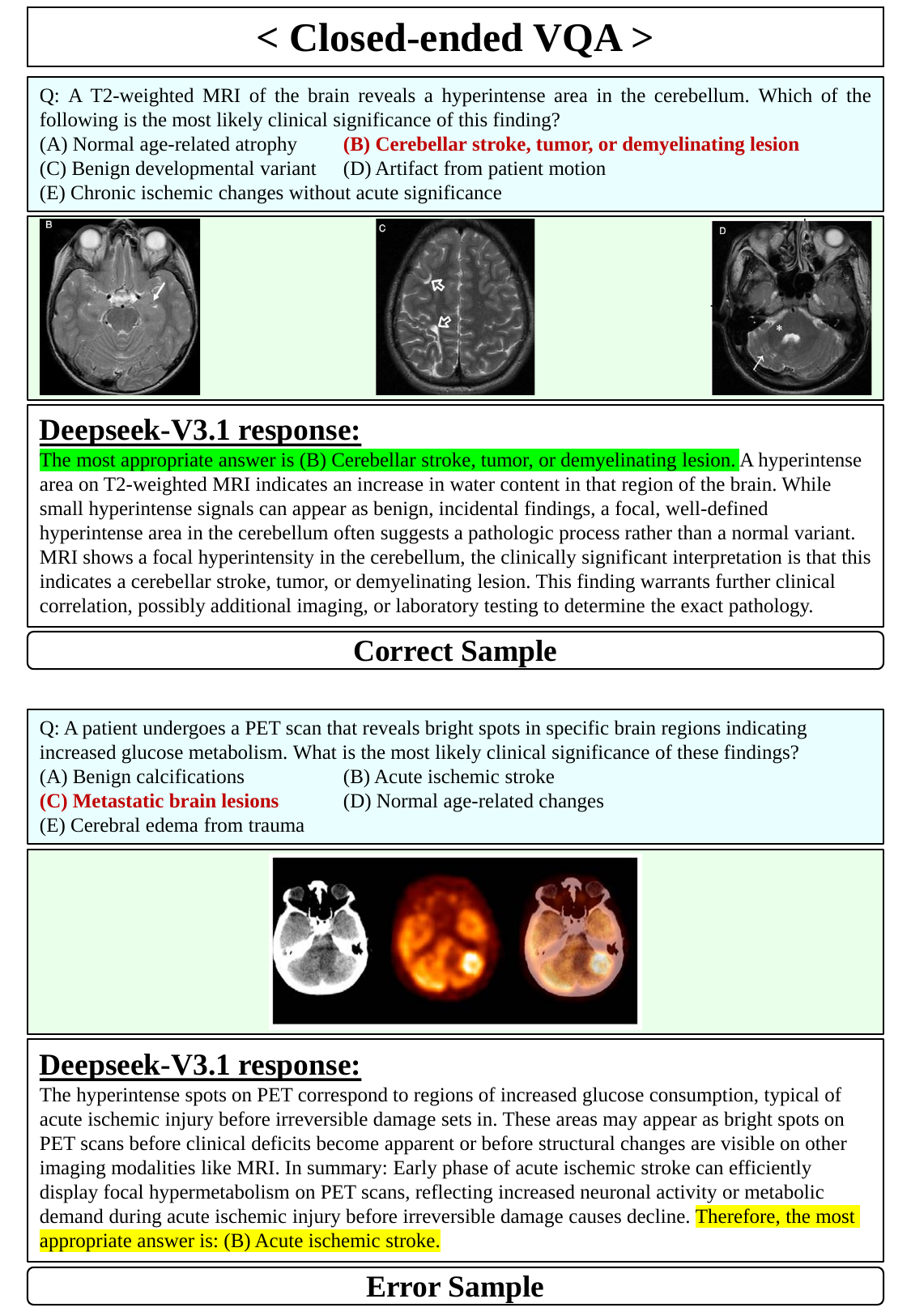}
   \caption{Correct/Error samples in Deepseek-V3.1 closed-ended VQA.}
   \label{fig: Deepseek-V3.1 closed-ended VQA}
\end{figure*}

\begin{figure*}[t]
  \centering
   \includegraphics[width=0.88\linewidth]{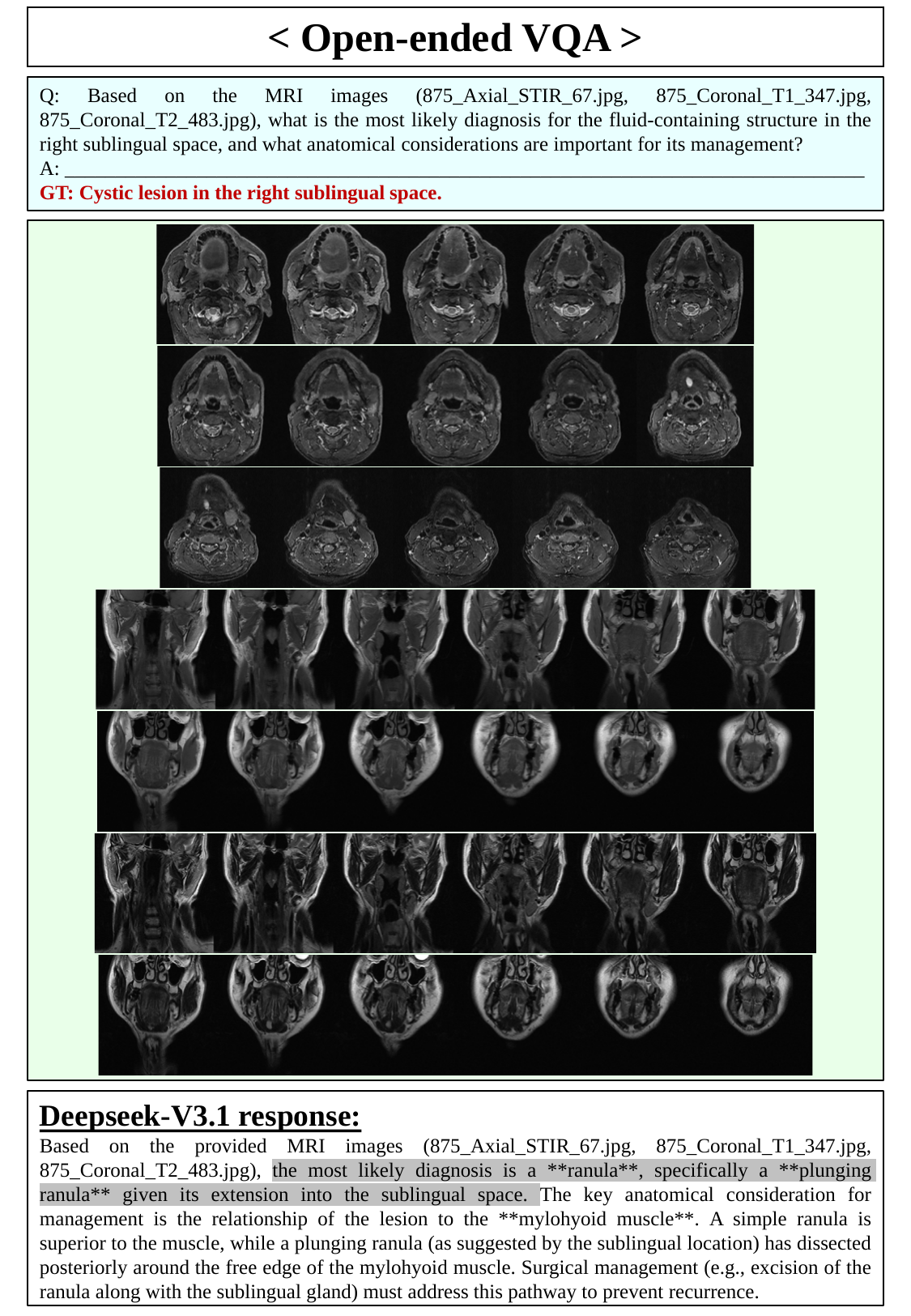}
   \caption{Correct/Error samples in Deepseek-V3.1 open-ended VQA.}
   \label{fig: Deepseek-V3.1 open-ended VQA}
\end{figure*}

\begin{figure*}[t]
  \centering
   \includegraphics[width=0.88\linewidth]{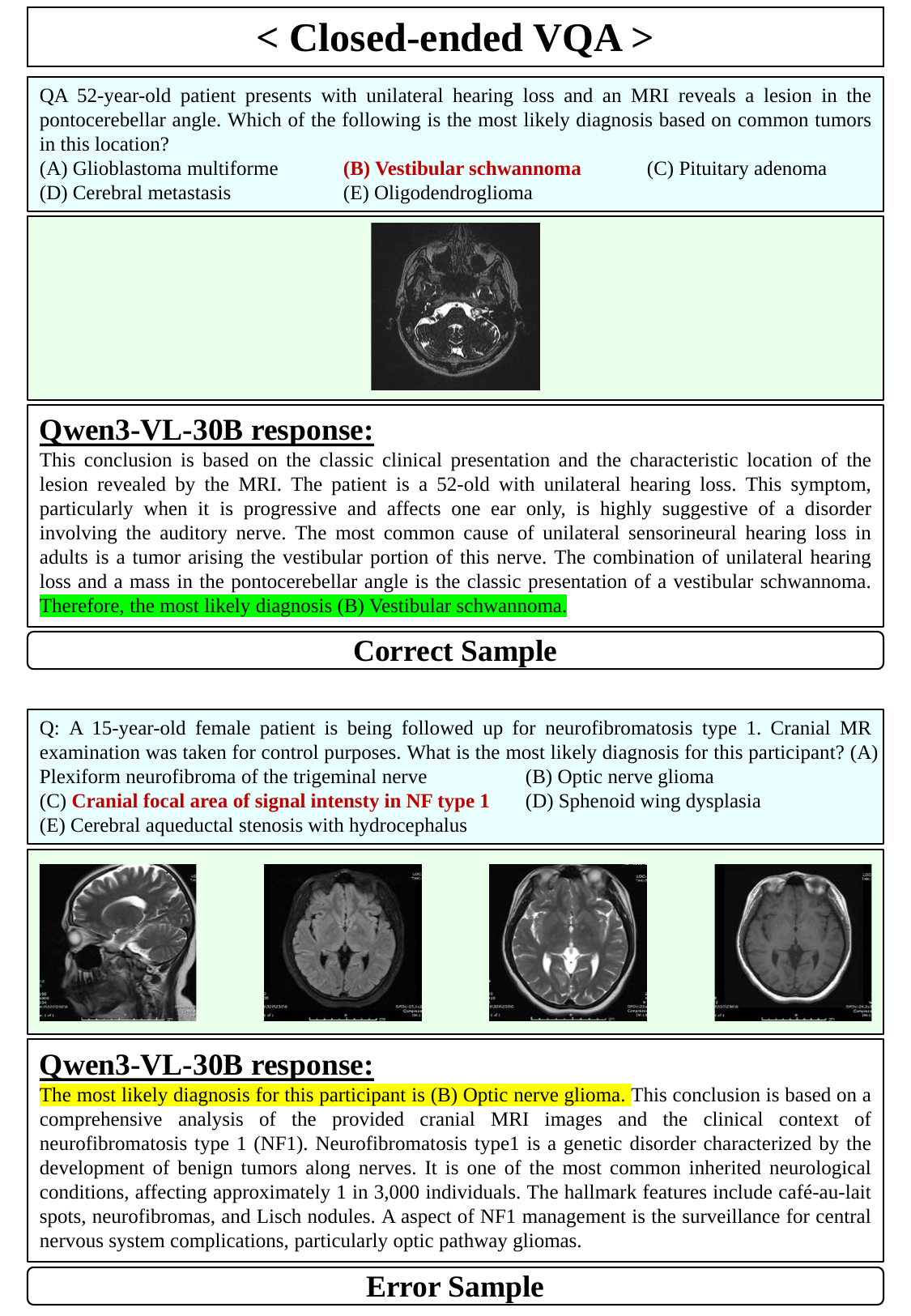}
   \caption{Correct/Error samples in Qwen3-VL-30B closed-ended VQA.}
   \label{fig: Qwen3-VL-30B closed-ended VQA}
\end{figure*}

\begin{figure*}[t]
  \centering
   \includegraphics[width=0.88\linewidth]{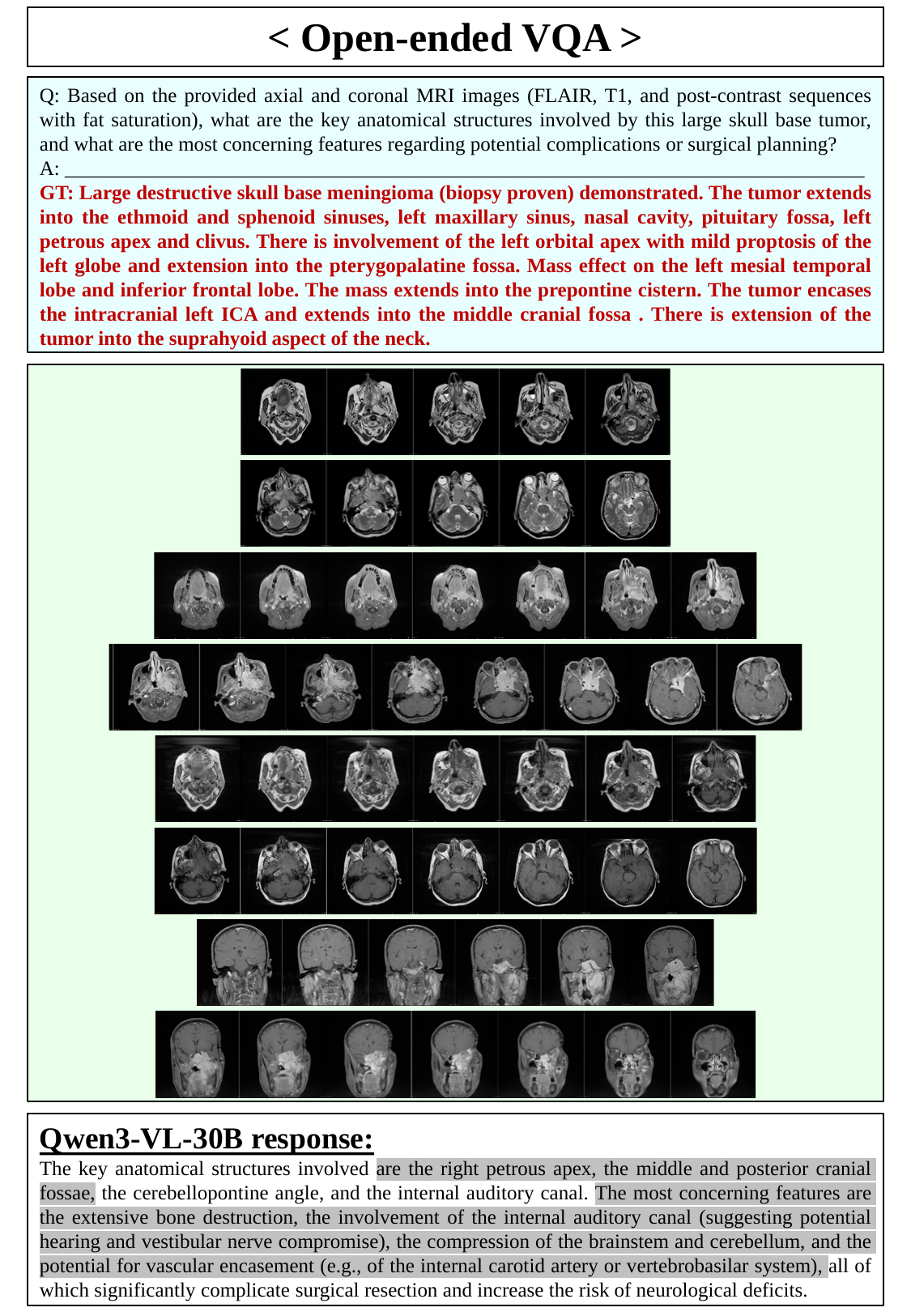}
   \caption{Correct/Error samples in Qwen3-VL-30B open-ended VQA.}
   \label{fig: Qwen3-VL-30B open-ended VQA}
\end{figure*}

\begin{figure*}[t]
  \centering
   \includegraphics[width=0.88\linewidth]{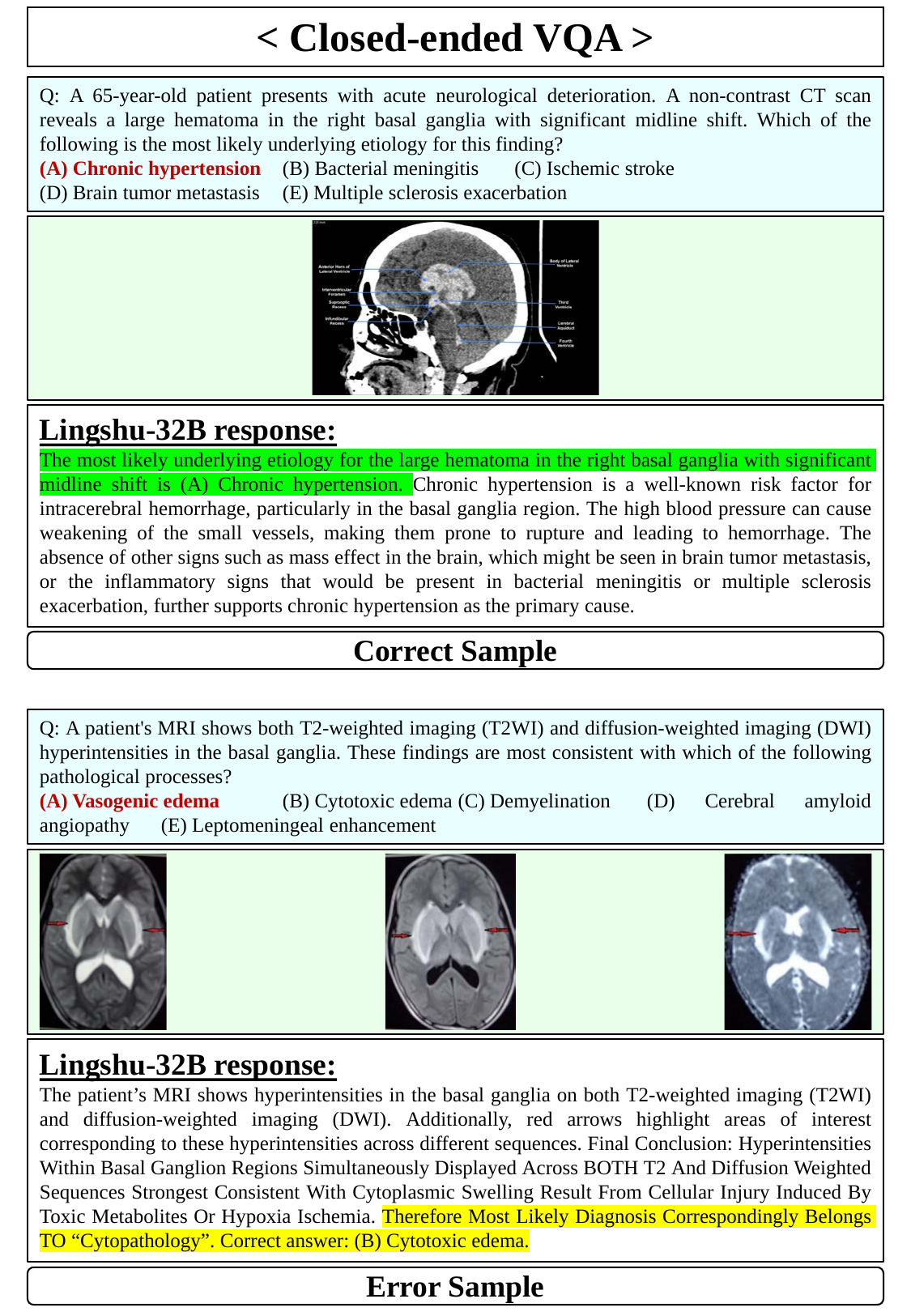}
   \caption{Correct/Error samples in Lingshu-32B closed-ended VQA.}
   \label{fig: Lingshu-32B closed-ended VQA}
\end{figure*}

\begin{figure*}[t]
  \centering
   \includegraphics[width=0.88\linewidth]{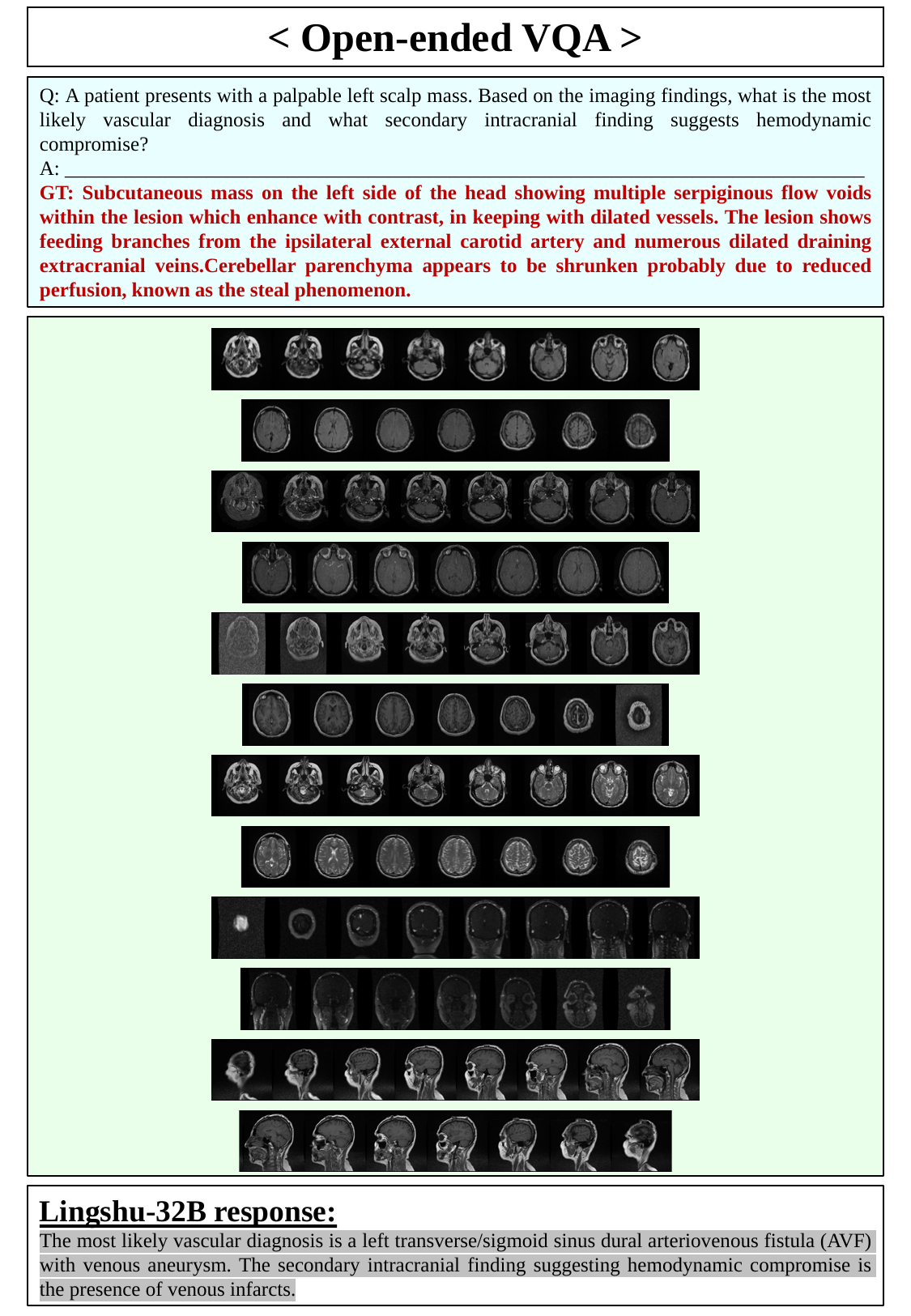}
   \caption{Correct/Error samples in Lingshu-32B open-ended VQA.}
   \label{fig: Lingshu-32B open-ended VQA}
\end{figure*}

\section{Case Study}
In this section, we conduct a comprehensive case study analysis of multiple MLLMs in our~\ourbench~under various scenarios. The evaluation is structured into two primary tracks: closed-ended VQA and open-ended VQA, allowing for a nuanced assessment of model capabilities across different task formats.

\textbf{Correct Samples.} From \cref{fig: GPT-5 closed-ended VQA,fig: GPT-5 open-ended VQA,fig: Claude-4.5-Sonnet closed-ended VQA,fig: Claude-4.5-Sonnet open-ended VQA,fig: Gemini-2.5-Pro closed-ended VQA,fig: Gemini-2.5-Pro open-ended VQA,fig: Deepseek-V3.1 closed-ended VQA,fig: Deepseek-V3.1 open-ended VQA,fig: Qwen3-VL-30B closed-ended VQA,fig: Qwen3-VL-30B open-ended VQA,fig: Lingshu-32B closed-ended VQA,fig: Lingshu-32B open-ended VQA}, our closed- and open-ended evaluations reveal that state-of-the-art models demonstrate a high degree of proficiency in both accurately interpreting brain imaging data and generating clinically actionable insights. The prevalence of open-ended VQA instances further underscores the models' strong performance in generating detailed, free-form explanations, which is critical for comprehensive diagnostic support. These capabilities indicate a promising role for such models in assisting real-world brain imaging analysis.

\textbf{Error Case Analysis.} A fine-grained analysis of these errors reveals three predominant failure modes:

\begin{itemize}
    \item \textbf{Perception Error.} It occurs when the model fails to correctly identify or localize fundamental visual elements within the brain scan, where a MLLM might misidentify a specific brain structure, overlook a small lesion, or perceive the boundaries of an anomaly. For example, in Fig. \ref{fig: Gemini-2.5-Pro closed-ended VQA}, Gemini-2.5-Pro \cite{comanici2025gemini} likely suffered from a subtle perception error. The MLLM may have correctly perceived the bilateral thalamic hyperintensities but failed to correctly perceive or localize the specific vascular territory involved on the imaging slices. In addition, in Fig. \ref{fig: Lingshu-32B closed-ended VQA}, Lingshu-32B \cite{xu2025lingshu} demonstrates a critical perception error by failing to observe the fundamental imaging finding of an absent apparent diffusion coefficient signal, which is essential for confirming the true nature of the DWI hyperintensity and making a correct diagnosis.
    \item \textbf{Understanding Error.} The model accurately perceives the visual features but fails to grasp their clinical significance or context, confusing one type of lesion for another or failing to link a radiographic finding to a potential pathology. For example, in Fig. \ref{fig: GPT-5 closed-ended VQA}, GPT-5 \cite{openai2025gpt5} demonstrates a clear understanding error by failing to grasp the critical epidemiological context that cerebellar metastases are the most common cerebellar tumor, while hemangioblastomas are rare in this age group. Moreover, in Fig. \ref{fig: Deepseek-V3.1 closed-ended VQA}, Deepseek-V3.1 \cite{guo2025deepseek} demonstrates a fundamental understanding error by failing to grasp the basic clinical significance of increased glucose metabolism on a PET scan, constructing its explanation around a pathophysiologically implausible and factually incorrect premise.
    \item \textbf{Reasoning Error.} The model may correctly perceive and understand individual elements but then make an incorrect clinical deduction. The cases often fall, where the reasoning process of the MLLM becomes opaque or logically inconsistent, resulting in nonsensical or unjustified conclusions. For example, in Fig. \ref{fig: Claude-4.5-Sonnet closed-ended VQA}, Claude-4.5-Sonnet \cite{anthropic2025claude45} demonstrates a reasoning error by constructing a detailed, post-hoc justification for a decision-making that is epidemiologically improbable, while completely failing to consider the most likely cause of adult-onset focal dystonia. Moreover, in Fig. \ref{fig: Qwen3-VL-30B closed-ended VQA}, Qwen3-VL-30B \cite{qwen3-VL} demonstrates a classic reasoning error by substituting general textbook knowledge for a specific brain imaging analysis of the provided options, leading to a conclusion that is logically disconnected from the most probable and contextually appropriate answer.
\end{itemize}

These findings emphasize that while the leading models are highly capable, their deployment in sensitive medical contexts requires careful validation and further refinement to mitigate these specific error types and ensure consistent, interpretable, and reliable results.

\begin{table*}[]
\caption{Diverse disease coverage on our~\ourbench.}
\label{tab:diverse disease}
\resizebox{0.92\linewidth}{!}{
\begin{tabular}{lll}
\hline\hline
\multicolumn{3}{c}{Diverse Disease Coverage}                                                                                                                       \\ \hline
Tumor                                  & Toxic Or Metabolic Encephalopathies           & Metastatic Brain Lesions                                 \\
Stroke                                 & Granulomatosis With Polyangiitis              & Pericallosal Lipoma                                      \\
Aneurysm                               & Leptomeningeal Spread                         & Moyamoya Disease                                         \\
Meningioma                             & Vasculopathy w. Cerebral Leukoencephalopathy  & Fragile X-associated Tremor/ataxia Syndrome              \\
Glioma                                 & Familial Cerebral Cavernous Malformation      & Cavernous Malformation                                   \\
Glioblastoma                           & Cerebral Venous Air Embolism                  & Maxillary Sinusitis                                      \\
Pituitary Adenoma                      & Hypothalamic Hamartoma                        & Chordoma                                                 \\
Metastasis                             & Dolichoectasia                                & Sinus Pericranii                                         \\
Adenoma                                & Vexas Syndrome                                & Pontine Tuberculoma                                      \\
Astrocytoma                            & Arachnoid Cyst                                & Lewy Body Dementia                                       \\
Schwannoma                             & Cns Toxoplasmosis                             & Alpha-methylacyl-coa racemase deficiency                 \\
Arteriovenous Malformation             & Hypomelanosis Of Ito                          & Pineoblastoma                                            \\
Venous Sinus Thrombosis                & Methylmalonic Acidaemia                       & Epidural Hematoma                                        \\
Neoplasm                               & Cerebral Fat Embolism                         & Pilocytic Astrocytoma                                    \\
Multiple Sclerosis                     & Intracranial Atherosclerosis                  & Oropouche Virus Encephalitis                             \\
Temporal Lobe Epilepsy                 & Olfactory Bulb Meningioma                     & Mature Teratoma                                          \\
Cavernous Angioma                      & Uremic Encephalopathy                         & Neurosyphilis                                            \\
Parkinson's Disease                    & Adamantinomatous Craniopharyngioma            & Wilson Disease                                           \\
Lymphoma                               & Small Vessel Ischemic Disease                 & Cavernous Cerebral Malformation                          \\
Fahr's Syndrome                        & Lipoma Of The Corpus Callosum                 & Sphenoid Mucocele                                        \\
Tuberous Sclerosis Complex             & Mitochondrial Echs1 Deficiency                & Inflammatory Demyelinating Lesions                       \\
Epilepsy                               & Myeloid Sarcoma                               & Anterior Cerebral Artery Aneurysm                        \\
Vestibular Schwannoma                  & Vertebral Artery Dissection                   & Vermis Agenesis                                          \\
Subdural Hematoma                      & Demyelinating Disease                         & Cerebellar tumor                                         \\
Malignant Neoplasm                     & Frontal Sinusitis                             & Infratemporal Fossa Tumor                                \\
Wilson's Disease                       & Venolymphatic Malformation                    & Actinomyces Osteomyelitis                                \\
Vascular Malformation                  & Meningoencephalitis                           & Clivus chordoma                                          \\
Hemangioblastoma                       & Lissencephaly                                 & Small Vessel Disease                                     \\
Focal Cortical Dysplasia               & Reversible Cerebral Vasoconstriction Syndrome & Central Pontine Myelinolysis                             \\
Epidermoid                             & Late-infantile Metachromatic Leukodystrophy   & Angiosarcoma                                             \\
Acoustic Neuroma                       & HSV Encephalitis                              & Carotid Artery Dissection                                \\
Joubert Syndrome                       & Venous Malformation                           & Baló's concentric sclerosis                              \\
Alzheimer's Disease                    & Frontotemporal dementia                       & Progressive Supranuclear Palsy                           \\
Hypertrophic Olivary Degeneration      & Mild Encephalitis/encephalopathy              & Megalencephaly                                           \\
Frontotemporal Dementia                & Haemangioblastoma                             & Anterior Cerebral Artery Stroke                          \\
Hypopituitarism                        & Disorder Of Glycosylation Type-1a             & Jugular Bulb Thrombosis                                  \\
Pontocerebellar Hypoplasia             & Midbrain Tectum Glioma                        & Pachygyria                                               \\
Craniosynostosis                       & H3K27M-mutant glioma                          & Poretti-boltshauser Syndrome                             \\
Hemimegalencephaly                     & Dyke-Davidoff-Masson Syndrome                 & Human Immunodeficiency Virus Dementia                    \\
Wernicke’s Encephalopathy              & Agenesis of the Septum Pellucidum             & Lipoma                                                   \\
Choroid Plexus Papilloma               & Benign Or Low-grade Neoplasm                  & Lgi1 Autoimmune Encephalitis                             \\
Brainstem glioma                       & Atherosclerosis                               & Myelin Oligodendrocyte Glycoprotein Antibody             \\
Central Neurocytoma                    & Vermis hypoplasia                             & Primary Angiitis Of The Central Nervous System           \\
Central Nervous System Germinoma       & Posterior Cortical Atrophy                    & Methanol Toxicity                                        \\
Transient Ischemic Attack              & Encephalitis                                  & Anterior Choroidal \& Thalamoperforate Arteries Syndrome \\
Rhinocerebral Mucormycosis             & Juvenile Angiofibroma                         & Orbital Cellulitis                                       \\
Chiari I Malformation                  & Fibromuscular Dysplasia                       & Corpus Callosum Agenesis                                 \\
Dysplastic Cerebellar Gangliocytoma    & Dysplastic Gangliocytoma                      & Reversible Posterior Leukoencephalopathy Syndrome        \\
Paranasal Sinus Tumor                  & Congenital CMV Infection                      & Idiopathic Hyperatrophic Pachymeningitis                 \\
Enterovirus A71 Rhombencephalomyelitis & Rhino-orbital-cerebral Mucormycosis           & Vestibulocochlear Nerve Schwannoma                       \\
Anti-lg1 Receptor Encephalitis         & Carotid-cavernous Fistula                     & Fabry disease                                            \\
Textiloma                              & Toxoplasmosis                                 & Acute Complete Occlusion Of Internal Carotid Artery      \\
Limbic Encephalitis                    & Chronic Subdural Hematoma                     & Polymicrogyria                                           \\
Gliosarcoma                            & Plasmacytoma                                  & Cryptococcosis                                           \\
Marchiafava-bignami Disease            & Acute Subdural Hematoma                       & Medulloblastoma                                          \\
Progressive External Ophthalmoplegia   & Rhabdomyosarcoma                              & Chiari malformation type III                             \\
Herpes Simplex Encephalitis            & Posterior Reversible Encephalopathy Syndrome  & Acute Disseminated Encephalomyelitis                     \\
Hypoxic Ischemic Encephalopathy        & Nocardia Asteroides Infection                 & Congenital Fusion Of The Radius And Ulna                 \\
Meningomyelocele                       & Acute Necrotizing Encephalitis Of Childhood   & Cerebral Hydatid Disease                                 \\
Rabies                                 & Tuberculomas                                  & X-linked Adrenoleukodystrophy                            \\
Krabbes Disease                        & Hypoglycaemic Encephalopathy                  & Intraventricular Migration Of Intra-ocular Silicone Oil  \\
Leigh's Disease                        & HHV-6 Encephalitis                            & Autoimmune Subacute Encephalitis                         \\
Ependymoma                             & Basilar Artery Thrombosis                     & Ethmoid Sinusitis                                        \\
Ischemic Stroke                        & Meningioangiomatosis                          & Beta-propeller Protein-associated Neurodegeneration      \\
Vertebral artery dissection            & Neurodegeneration w. Brain Iron Accumulation  & Sphenoid Sinus Mucocele                                  \\
Arteriovenous Fistula                  & Prolactinoma                                  & Vein of Galen malformation                               \\
Cerebrotendinous Xanthomatosis         & Thrombosis Of The Dural Sinuses               & Agenesis Of The Corpus Callosum                          \\
Chronic Cerebrovascular Disease        & Transverse Sinus Thrombosis                   & Granulomatous Amebic Encephalitis                        \\
Bacterial Meningitis                   & Maxillary Sinus Tumor                         & Huntington's Disease                                     \\
Linear Scleroderma                     & Diabetic Striopathy                           & Low-grade Fibromyxoid Sarcoma                            \\
High-grade Neoplasm                    & Craniopharyngioma                             & Apert Syndrome                                           \\
Amyotrophic Lateral Sclerosis          & Vitamin B1 (thiamine) Deficiency              & Pericallosal Artery Aneurysm                             \\
Colpocephaly                           & Cerebral Air Embolism                         &                                                          \\ \hline\hline
\end{tabular}
}
\end{table*}



\end{document}